\documentclass[acmlarge]{acmart}
\usepackage{multirow}
\usepackage[normalem]{ulem}
\useunder{\uline}{\ul}{}
\usepackage{algorithm}
\usepackage{algorithmic}
\usepackage{xcolor}
\usepackage{xspace}
\usepackage{booktabs}
\usepackage{lineno,hyperref}
\usepackage{graphicx}
\usepackage{subfigure}

\usepackage{color}
\usepackage{amsmath}
\usepackage{graphicx}
\usepackage{subfigure}
\usepackage{diagbox}
\usepackage{subfloat}
\usepackage{verbatimbox}
\usepackage{float}
\usepackage{array}
\usepackage{booktabs}
\usepackage{threeparttable}
\usepackage{subcaption}
\usepackage{todonotes}
\usepackage[bottom]{footmisc}
\usepackage{amssymb}



\newcommand{\method}{OFTTA\xspace}
\AtBeginDocument{%
  }

\setcopyright{acmcopyright}
\copyrightyear{2023}
\acmYear{2023}
\acmDOI{XXXXXXX.XXXXXXX}

\acmJournal{IMWUT}
\acmVolume{37}
\acmNumber{4}
\acmArticle{111}
\acmMonth{8}

\begin{document}

\setcopyright{acmlicensed}
\acmJournal{IMWUT}
\acmYear{2023} \acmVolume{7} \acmNumber{4} \acmArticle{183} \acmMonth{12} \acmPrice{15.00}\acmDOI{10.1145/3631450}

\title{Optimization-Free Test-Time Adaptation for Cross-Person Activity Recognition}

\author{Shuoyuan Wang}
\email{claytonwang0205@gmail.com}
\orcid{0000-0003-1795-4161}
\affiliation{
  \department{PAMI Research Group, Department of Computer and Information Science}
  \institution{University of Macau}
  \city{Taipa}
  \country{Macau}
}

\author{Jindong Wang}
\email{jindong.wang@microsoft.com}
\orcid{0000-0002-4833-0880}
\affiliation{%
  \institution{Microsoft Research Asia}
  \city{Beijing}
  \country{China}
}

\author{Huajun Xi}
\orcid{0009-0005-2788-1121}
\email{12112806@mail.sustech.edu.cn}
\affiliation{%
  \institution{Southern University of Science and Technology}
  \city{Shenzhen}
  \state{Guang Dong}
  \country{China}
}

\author{Bob Zhang}
\email{bobzhang@um.edu.mo}
\orcid{0000-0003-2497-9519}
\affiliation{%
  \department{PAMI Research Group, Department of Computer and Information Science, University of Macau, Taipa, Macau, Centre for Artificial Intelligence and Robotics, Institute of Collaborative Innovation}
  \institution{University of Macau}
  \city{Taipa}
  \country{Macau}}
  \authornote{Corresponding author: Bob Zhang}

\author{Lei Zhang}
\email{leizhang@njnu.edu.cn}
\orcid{0000-0001-8749-7459}
\affiliation{%
 \institution{Nanjing Normal University}
 \city{Naning}
 \state{Jiang Su}
 \country{China}}

\author{Hongxin Wei}
\email{weihx@sustech.edu.cn}
\orcid{0000-0002-8973-2843}
\affiliation{%
  \department{Department of Statistics and Data Science}
  \institution{Southern University of Science and Technology}
  \city{Shenzhen}
  \state{Guang Dong}
  \country{China}
}

\renewcommand{\shortauthors}{Wang et al.}
\begin{abstract}

Human Activity Recognition (HAR) models often suffer from performance degradation in real-world applications due to distribution shifts in activity patterns across individuals. Test-Time Adaptation (TTA) is an emerging learning paradigm that aims to utilize the test stream to adjust predictions in real-time inference, which has not been explored in HAR before. However, the high computational cost of optimization-based TTA algorithms makes it intractable to run on resource-constrained edge devices. In this paper, we propose an \textbf{O}ptimization-\textbf{F}ree \textbf{T}est-\textbf{T}ime \textbf{A}daptation (OFTTA) framework for sensor-based HAR. OFTTA adjusts the feature extractor and linear classifier simultaneously in an \emph{optimization-free} manner. For the feature extractor, we propose \textbf{E}xponential \textbf{D}ecay \textbf{T}est-time \textbf{N}ormalization (EDTN) to replace the conventional batch normalization (CBN) layers. EDTN combines CBN and Test-time batch Normalization (TBN) to extract reliable features against domain shifts with TBN's influence decreasing exponentially in deeper layers. For the classifier, we adjust the prediction by computing the distance between the feature and the prototype, which is calculated by a maintained support set. In addition, the update of the support set is based on the pseudo label, which can benefit from reliable features extracted by EDTN. Extensive experiments on three public cross-person HAR datasets and two different TTA settings demonstrate that OFTTA outperforms the state-of-the-art TTA approaches in both classification performance and computational efficiency. Finally, we verify the superiority of our proposed OFTTA on edge devices, indicating possible deployment in real applications. 

\end{abstract}





\maketitle

\section{Introduction}
\indent Sensor-based human activity recognition (HAR) is an active research field in ubiquitous and mobile computing. The main goal of sensor-based HAR is to build machine learning models to recognize various human activities accurately and efficiently. Recently, with the rapid advances in sensor technology and IoT-enabled systems, HAR has gained numerous applications such as healthcare monitoring \cite{wang2019survey}, smart homes \cite{bianchi2019iot}, sports analytics \cite{pareek2021survey}, etc. The main challenge in sensor-based HAR is to process multi-variate sensor data and predict human activities accurately and efficiently. To solve this problem, researchers have employed machine learning (ML) methods \cite{khemchandani2016robust} and deep learning (DL) methods \cite{yang2015deep, grushin2013robust} to boost HAR performance.

\begin{figure*}[t!]
  \centering
  \begin{minipage}[t]{0.30\textwidth}
    \centering
    \includegraphics[width=\textwidth]{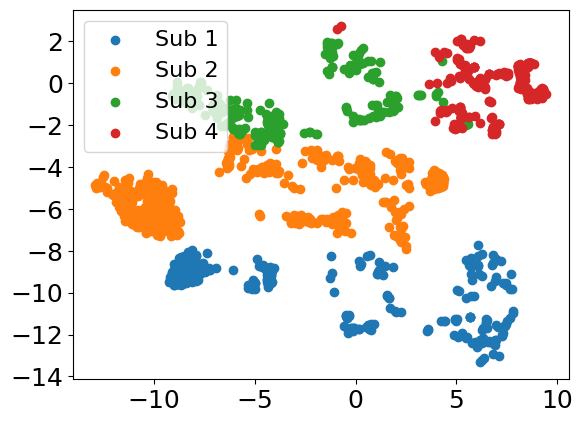}
    \caption{T-SNE visualization in the domain space using UniMiB. Each domain is denoted by a specific color.}
    \label{fig:1}
  \end{minipage}
  \hspace{0.3cm}
  \begin{minipage}[t]{0.29\textwidth}
    \centering
    \includegraphics[width=\textwidth]{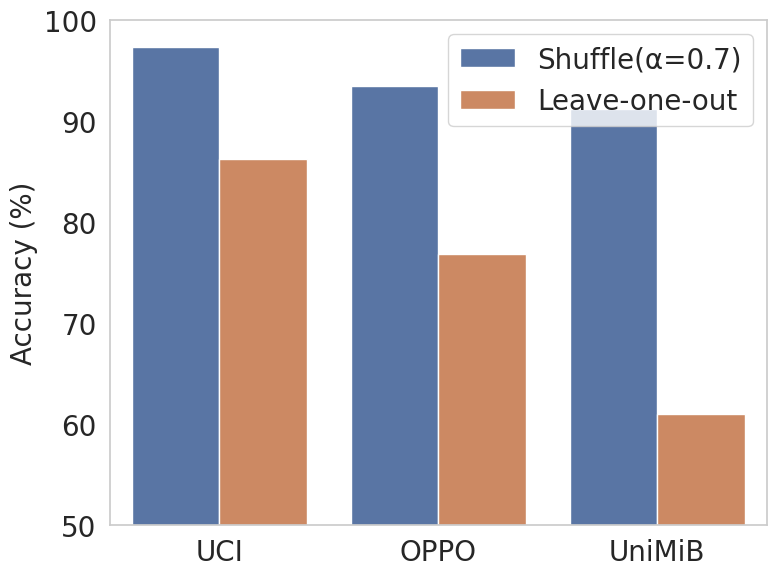}
    \caption{Classification comparison between random splitting and the subject-wise splitting strategy.}
    \label{fig:2}
  \end{minipage}
  \hspace{0.3cm}
  \begin{minipage}[t]{0.335\textwidth}
    \centering
    \includegraphics[width=\textwidth]{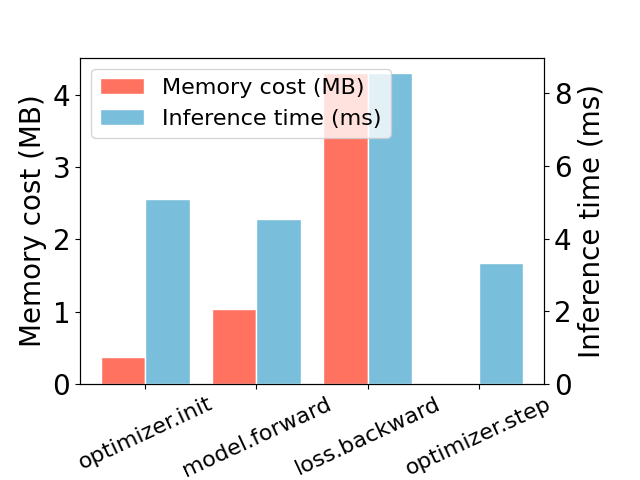}
    \caption{Incremental memory and time cost brought by TENT optimization.}
    \label{fig:3}
  \end{minipage}
\end{figure*}

However, due to inter-subject variability including age, body shape, behavioral habits, or other factors in individual activity patterns, the HAR model may not always perform well.
One of the main challenging scenarios is when applied to a \emph{new} test environment, HAR models may generalize poorly if the data come from \emph{different} distributions from the training data. 
Hence, if a HAR model is directly applied to a distribution (e.g., person), model performance may deteriorate and the failure leads to potential risk due to the uncertainty caused by \emph{distribution shifts}. An intuitive visualization is shown in Figure \ref{fig:1}, we use t-SNE \cite{van2008visualizing} to visualize the features from the UniMiB dataset. The features extracted from each subject are denoted by a specific color. Assuming we follow the leave-one-out rule to use one subject data as the testing set, it is clear that the testing data distributions almost do not overlap with any of the existing source domains. Instead, they are distributed over the unfilled domain space, indicating exploration of unseen domains. To further verify the potential risk of distribution shift, we compare the impact of two dataset splitting strategies on classification accuracy in Figure \ref{fig:2}. “shuffle” means the dataset is randomly split into the training set and the testing set with a ratio of 7:3, which is a traditional splitting method. “Leave-one-out” is the person-wise dataset splitting strategy in our experiment, and we report the domain-wise average classification results. It is clear that the classification performance will noticeably decrease if the model can not access the data from the specific domain, which is a prevalent problem in real life where it is difficult to collect sufficient user data and annotate them in reality.

To overcome distribution shifts in HAR, one intuitive solution is to collect massive sensor data. However, in real applications, it is time-consuming and impossible to collect all personalized data \cite{su2022learning}.
In recent years, distribution shift has gained increasing attention, and techniques like domain adaptation (DA) \cite{wang2019deep} and domain generalization (DG) \cite{wang2022generalizing}. Those techniques have been successfully applied to mitigate the domain shift in HAR scenarios \cite{chang2020systematic, lu2022semantic, lu2022local, qin2022domain}.
However, one main drawback is DA needs to access data from both the source domain and target domain, which may cause privacy and security concerns \cite{bourtoule2021machine}.
While DA incorporates target domain data into the training, DG only needs data from one or several domains and expects the model to generalize to the unseen domain.  
However, recent research shows that it may be infeasible to generalize a model well to unseen distributions if the model can not utilize the target data \cite{zhang2022domain}. As an achievable alternative, we can loose the restriction and learn knowledge sequentially from the target domain in an online unsupervised manner during inference time, which is an intuitive explanation of Test-Time Adaptation (TTA) \cite{liang2023comprehensive, wang2020tent}.
Compared with DG, TTA can utilize knowledge from the target domain and adjust the model in an online manner.
Moreover, compared with DA, TTA does not incorporate personal activity information into the training phase and can benefit privacy protection. As a novel learning paradigm, TTA provides a feasible solution for HAR against distribution shifts.

Considering recent advances in TTA, \textbf{can we directly apply existing TTA methods to sensor-based HAR?}
Unfortunately, this will face multiple challenges.
\emph{First}, Most HAR systems are deployed on resource-constrained edge devices with limited computational resources.
However, most state-of-the-art TTA techniques require parameter optimization \cite{wang2022continual, wang2020tent, niu2023towards}. Unfortunately, this adaptation paradigm may bring two potential risks: (1) \textbf{Overhead computation}: an advanced HAR model in real-world scenarios needs to adapt to diverse distributions accurately and efficiently. Gradient updates may bring heavier computational burden and hinder real-time inference. For instance, we use TENT \cite{wang2020tent} to adapt the model on the UCIHAR dataset \cite{anguita2012human} in Figure \ref{fig:3}. Compared to model inference (\textit{model.forward}), gradient optimization (\textit{loss.backward}, \textit{optimizer.step}) brings a larger cache size and slower inference speed, which hinders HAR in real applications. (2) \textbf{Catastrophic forgetting}:  while existing TTA solutions can significantly improve the classification performance on out-of-distribution data, some test samples may bring noisy gradients. As a result, the adapted model may be disrupted and gradually lose its prediction ability on samples from the training domains. Hence, using optimization-free adaptation can be a feasible solution. To date, some researchers have explored efficient TTA without updating parameters. T3A \cite{iwasawa2021test} simply adjusts the linear classifier via a prototype-based method. The prototype is computed by the centroid of the class-wise support set they proposed. AdaNPC \cite{zhang2023adanpc} replaces the linear layer with a K-Nearest Neighbors (KNN) classifier. 
Both works adopt pseudo labels as the criterion to update the memory bank or the support set. However, without supervised information, the pseudo label may not always be reliable, and even the model has high confidence in the data \cite{wu2023uncovering}. If we continually update the memory bank by noisy pseudo label, the memory bank will be gradually polluted and generate biased centroid or visit evil neighbors. 
Therefore, simple classifier adjustment may not be optimal. In short, we have two challenges: \textbf{(1) How to conduct efficient and stable TTA in HAR?} and \textbf{(2) How to prevent updating the memory bank with unreliable pseudo labels?}

\begin{figure}[htbp]
  \centering
  \subfigure[CBN]{
    \includegraphics[width=0.25\linewidth]{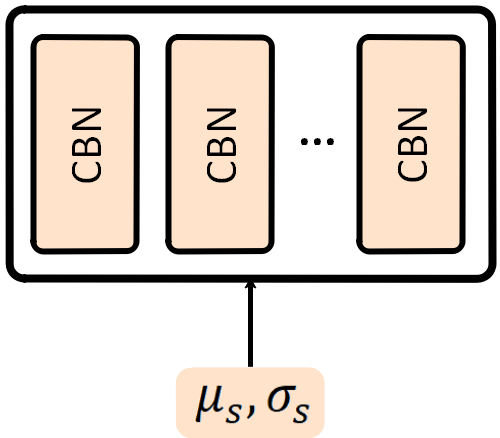}
    \label{fig:subfig4A}
  }
  \hspace{0.3cm}
  \subfigure[TBN]{
    \includegraphics[width=0.25\linewidth]{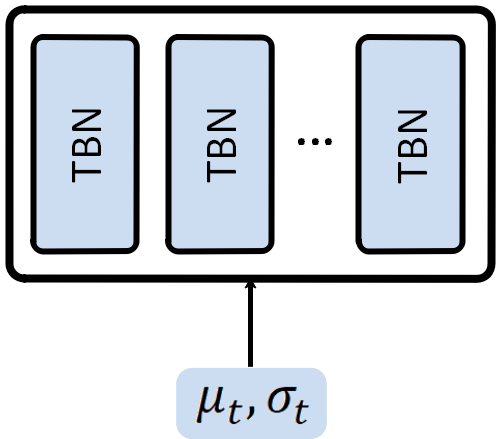}
    \label{fig:subfig4B}
  }
  \hspace{0.3cm}
  \subfigure[EDTN]{
    \includegraphics[width=0.25\linewidth]{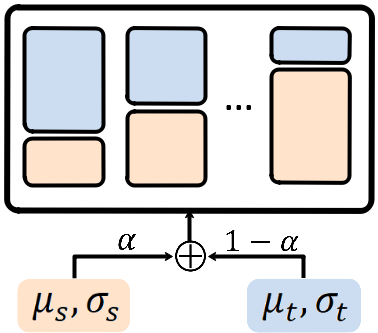}
    \label{fig:subfig4C}
  }
  \vspace{-.2in}
  \caption{The visualization of different batch statistics, where ${{\mu }_{s}}$ and ${{\sigma }_{s}}$ are the mean and variance of the source statistics. ${{\mu }_{t}}$ and ${{\sigma }_{t}}$ are the mean and variance of the statistics of the test batch. Our EDTN combines TBN and CBN in an exponential decay manner, where CBN accounts for a larger proportion in the deep layer.}
  \vspace{-.1in}
  \label{fig:4}
\end{figure}

To tackle the above issues, we propose an \textbf{O}ptimization-\textbf{F}ree \textbf{T}est-\textbf{T}ime \textbf{A}daptation framework for sensor-based HAR, short as \textbf{OFTTA}. As shown in Figure \ref{fig:5},
the key of OFTTA is to adjust the feature extractor and the classifier simultaneously \emph{without} parameter optimization.
Specifically, for the feature extractor, we proposed an \textbf{E}xponential \textbf{D}ecay \textbf{T}est-time \textbf{N}ormalization (\textbf{EDTN}) module to replace the conventional batch normalization (CBN) layers in the model pre-trained on the source domain. 
An illustrative visualization is shown in Figure \ref{fig:4} to help the readers understand the differences between CBN, TBN and EDTN quickly.
Limited by receptive field, test-time batch normalization (TBN) computes statistics on the current mini-batch and standardizes the local feature in the shallow layer, which is sensitive to style. In contrast, as the layer goes deeper, CBN shows its strength since features are extracted from a global scale, including more content information. Since content information like activity category information has been well-supervised and trained on the source domain, the parameters are not sensitive to distribution shifts and we can use more CBN in the deep layer. In other words, TBN can better standardize local features in shallow layers under distribution shifts, while CBN is more reliable in the deep layer.
EDTN mixes CBN and TBN, using more CBN than TBN when the layer goes deeper. This strategy enables them to work in a trade-off relationship and extract reliable features against distribution shifts. 
For the linear classifier, we simply adjust the prediction based on the pseudo-prototype. In particular, we utilize a support set to memory features by class. The support set is updated via the pseudo label and entropy ranking.
The prototype is adjusted by the centroid of each class, which is computed by low-entropy features in the support set. Finally, we obtain the adjusted prediction based on the distance between features and the nearest prototype. We empirically show that the feature extractor and classifier can be adjusted separately and work collaboratively. EDTN can help the model extract stable features and further compute more reliable pseudo labels. Note that OFTTA works in an optimization-free manner and does not alter the training phase, which is more preferable for HAR tasks on edge devices. 

To sum up, the main contribution of this paper is as follows:
\begin{enumerate}
    \item \textbf{New research direction}: We indicate a new research direction for generalizable HAR. We first explore test-time adaptation for sensor-based human activity recognition. Compared with other transfer learning methods, TTA is more achievable against domain shifts in real-world HAR.
    \item \textbf{Novel perspective}: To achieve better feature representation under distribution shifts, we adjust the normalization by test-time batch statistics in a layer-wise manner. The adjusted features can further output more reliable pseudo labels and benefit the prototype-based classifier.
    \item \textbf{Novel methodology}: To achieve efficient TTA in HAR, we propose OFTTA to adjust the model in an optimization-free manner. The suggested framework is accurate, flexible, and hardware-friendly. 
    \item \textbf{Comprehensive evaluation}: Extensive experiments on three public cross-person HAR datasets and two different TTA settings demonstrate that our proposed OFTTA framework can significantly improve the model performance and outperform the state-of-the-art TTA approaches. In addition, the superiority of OFTTA is verified on edge devices.
\end{enumerate}

\section{Related Work}
\subsection{Human Activity Recognition}
With the proliferation of wearable devices and sensors,sensor-based HAR has become a hot-pot research field over the years. The main goal of HAR is to recognize human activities collected by multi-variate sensors using machine learning models \cite{ravi2005activity}. Early studies on HAR mostly relied on traditional machine learning techniques such as Support Vector Machines (SVM), and k-Nearest Neighbors (kNN) \cite{bao2004activity, khemchandani2016robust}. Limited by feature engineering, these traditional methods may not be able to capture the high-level sensor features required for accurate recognition. To overcome the limitations, researchers have utilized deep learning techniques to improve HAR performance in an end-to-end manner \cite{wang2019deep, zeng2014convolutional, ordonez2016deep}. However, traditional machine learning and deep learning methods assume that data are independent and identically distributed (i.i.d.) \cite{wang2022generalizing}. For data that does not conform to i.i.d. assumption states, which is common in real-life tasks, these data-driven methods may suffer from performance degradation. Recently, more attention has been paid to generalizable HAR, which is dedicated to maintaining competitive HAR performance under distribution drift \cite{qian2021latent, lu2022semantic, qin2022domain, lu2023out}. We here focus on cross-person activity recognition, which is one popular task in generalizable HAR. Some less familiar readers might still be curious that why each individual's data can be considered as a distinct domain, and there exists a domain gap between different domains. We emphasize the cross-person setting here. Recently works \cite{qian2021latent, lu2022semantic} show that if we split the dataset in a person-wise manner, the performance on the test dataset will plummet. Hence, cross-person HAR is a realistic setting of distribution shifts and we should actively tackle this problem for real-world HAR applications.

\subsection{Domain Adaptation and Generalization}
Domain adaptation (DA) and domain generalization (DG) are two related topics, which refer to transferring source knowledge to improve performance in the target domain.
To tackle domain shift, DA uses less annotated or no annotated target domain data in the training phase.
\cite{khan2018scaling} proposed HDCNN to adjust the weights adaptively from different distributions.
\cite{chang2020systematic} align the feature extracted from both the source domain and target domain to solve the wearing diversity issue. Due to the complexity of downstream tasks or privacy concerns, we may not have access to data from the target domain. As an alternative, DG has been proposed in recent years and learns a generalizable model on single or several source domains. Ideally, the prediction function is well-generalized and can work well on unknown domains. Under the setting of HAR, DG enables the model to generalize to new users or new sensor modalities without retraining.
\cite{qian2021latent} first utilizes DG in HAR and removes domain-specific representations via variational autoencoder (VAE).
\cite{lu2022semantic} combines semantic-aware Mixup and large margin loss to improve generalizable cross-domain Human Activity Recognition (HAR) by considering activity semantic ranges and enhancing discrimination to prevent misclassification caused by noisy virtual labels.
\cite{qin2022domain} proposed a unified framework that learns both domain-specific and domain-invariant representations.
Recently study shows that it may be over-optimistic for a model to generalize well to unseen distribution if the model can not utilize target data \cite{dubey2021adaptive, zhang2022learning}.
Here we focus on mitigating distribution shifts via adaptation in the testing phase. 

\subsection{Test-Time Adaptation}
Test-Time Adaptation (TTA) is a challenging and realistic setting, which aims to adapt models using mini-batch data in a single-pass manner \cite{liang2023comprehensive}. We emphasize the protocol of the TTA setting and the details can be summarized as follows: (1) TTA does not alter the training stage (e.g. training objective). (2) The training data is no longer accessible during the testing phase. (3) Test data will be only adapted in a sequential one-pass manner.
Here we introduce works that strictly conform to this protocol. As a pioneering work, TENT \cite{wang2020tent} optimizes the model for confidence by minimizing the entropy of its predictions and updates normalization statistics online on each batch. T3A \cite{iwasawa2021test} adjusts the last trained linear classifier by computing pseudo-prototype representations. To tackle the non-stationary adaptation issue, COTTA \cite{wang2022continual} improves TTA by using a weight-averaged teacher model and augmentation-averaged predictions to improve the quality of the pseudo label. Stochastic restoration is also proposed to avoid catastrophic forgetting. EATA \cite{niu2022efficient} excludes samples with high entropy or are very similar to reduce the optimization cache. For TTA in the wild, SAR \cite{niu2023towards} combines sharpness-aware minimization (SAM) and reliable entropy minimization to perform adaptation more stably. To solve the class-imbalance data stream in TTA, DELTA \cite{zhao2023delta} renormalizes BN statics and punishes the weight of the dominant class. For efficient TTA, MECTA \cite{hong2023mecta} stops the back-propagation conditionally and prunes the model. To our best knowledge, we explore TTA for sensor-based human activity recognition for the first time.

\subsection{Entropy Minimization}
Entropy Minimization is a popular optimization scheme for TTA, which is first utilized in TENT \cite{wang2020tent}. Entropy minimization reduces the uncertainty of the model via high-density decision punishment. MEMO \cite{zhang2022memo} optimizes entropy of average prediction from different augmentations. EATA \cite{niu2022efficient} removes high entropy samples to prevent the model from collapsing. DELTA \cite{zhao2023delta} utilizes a class-wise re-weighting strategy for entropy loss in class-imbalance TTA. These methods regularize entropy via explicit parameter optimization in testing. As we discussed in Figure \ref{fig:3}, parameter optimization brings intensive computation on memory-constrained devices. For efficient inference on edge devices, we propose an optimization-free TTA framework via implicit entropy minimization. 
 
\section{\method: OPTIMIZATION-FREE TEST-TIME ADAPTATION FOR HAR}
In this section, we present our OFTTA framework for efficient sensor-based human activity recognition.

\subsection{Problem Formulation}

Given a set of source domains ${{\mathcal{D}}^{S}}=\{{{D}^{1}},{{D}^{2}},\ldots ,{{D}^{n}}\}$, each domain ${{D}^{i}}$ consists of a set of instances $(\boldsymbol{x}_{s}, y_{s}) \sim \mathcal{P}^{i}(\boldsymbol{x}, y)$, where ${{\mathbf{x}}_{s}}\in \mathcal{X}\subset {{\mathbb{R}}^{d}}$ denotes the $d$-dimensional sensor input, ${{y}_{s}}\in \mathcal{Y}=\{1,...,K\}$ is the corresponding label in $K$ activity categories, and ${{\mathcal{P}}^{i}}(\boldsymbol{x},y)$ is the underlying joint distribution for domain $D^{i}$. In the testing phase, we have data from a previously \emph{unseen} target domain $D^{T}$ with a different distribution $\mathcal{P}^{T}$. Noted that ${{\mathcal{P}}^{i}}(\boldsymbol{x},y)\ne {{\mathcal{P}}^{j}}(\boldsymbol{x},y),\forall i,j\in \{1,2,...,S,T\}$. Empirically, we hope the model ${f}^{*}$ pre-trained on the source domain can minimize the prediction error $\epsilon_t$ on the target domain:
\begin{equation}
\epsilon_t = \mathbb{E}_{(\boldsymbol{x}, y) \sim \mathcal{P}^{T}(\boldsymbol{x}, y)}\mathcal{L} (f^*(\mathbf{x}), y),
\label{eq:1}
\end{equation}
where $\mathbb{E}$ denotes the expectation and $\mathcal{L}\left( \cdot  \right)$ is the loss function. Note that Eq.~\eqref{eq:1} is only used for evaluation due to the unavailability of the target distribution. However, if the model can not utilize the target data, the model can not generalize well to unseen distribution \cite{dubey2021adaptive, zhang2022learning}, and the expectation will not be zero. As a realistic alternative, we have test-time adaptation(TTA) to mitigate the domain gap between the source domain and the target domain. Since the target domain is split from the origin dataset in a subject-wise manner, the target data does not lead to label distribution shifts during testing. In other words, we are focusing on covariate shift problems, and both the source and target data are in the same label space.

\begin{definition}[Test-time adaptation for HAR]
    Given a trained human activity recognition model $\theta^0$ and a sequence of unlabeled mini-batches of activity data $\{{{x}^{{{b}_{1}}}},{{x}^{{{b}_{2}}}},...{{x}^{{{b}_{n}}}}\}\in {{\mathcal{D}}^{T}}$, test-time adaptation (TTA) for HAR aims at learning an optimal activity recognition model $\theta^*$ by leveraging $\theta_0$ in a sequential manner:
    \begin{equation}
        {{\theta }_{t}}\leftarrow \underset{{{\theta }_{t-1}}}{\mathop{\arg \min }}\,{{\mathcal{L}}_{\text{TTA}}}\left( {{f}_{t-1}}({{x}^{{{b}_{t}}}}),{{\theta }_{t-1}} \right),
        \label{eq:2}
    \end{equation}
    where ${{\mathcal{L}}_{\text{TTA}}}(\cdot )$ is a generic TTA optimization objective and we can use the adjusted model ${{f}_{{{\theta }_{t}}}}$ to modify the prediction ${{\hat{y}}_{t}}=f({{x}^{{{b}_{t}}}},{{\theta }_{t}})$ while storing the knowledge gained from previously encountered mini-batches. We aim to develop a TTA algorithm to minimize the risk on $D^{T}$ with the adjusted model in the testing phase.
\end{definition}

\begin{figure*}[htbp]
	\hspace*{0cm}
	\centering
	\includegraphics[scale=0.16]{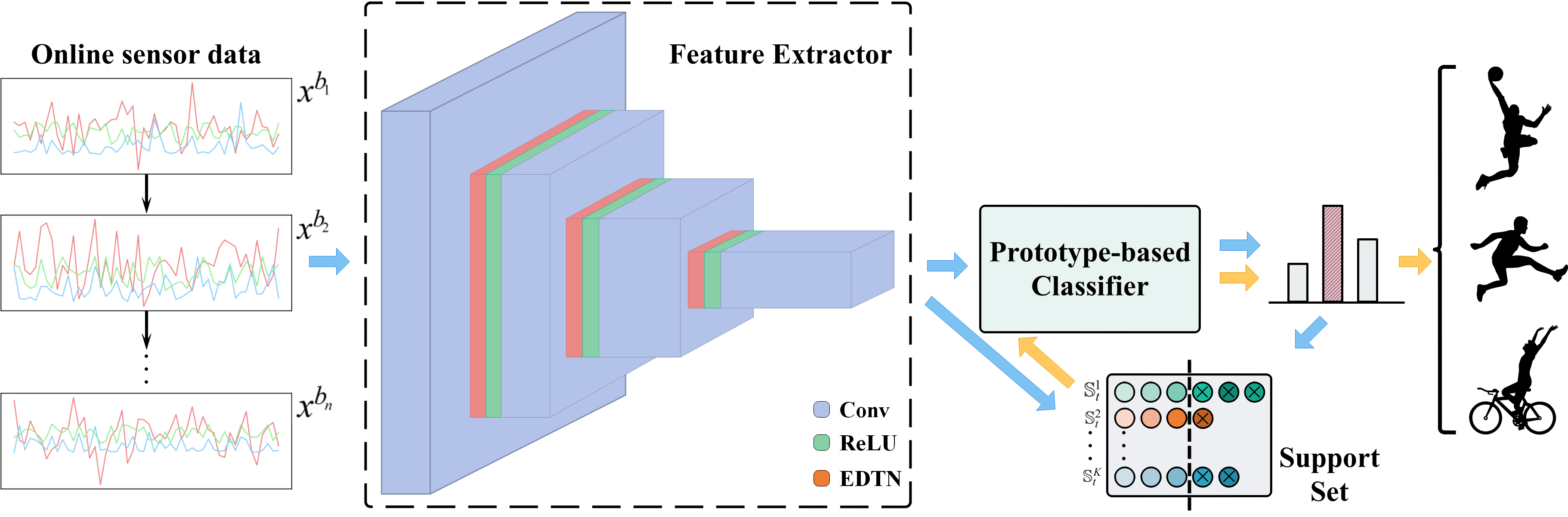}
	\caption{\label{Fig:main} \normalsize{Overview of the proposed OFTTA framework.}}
\label{fig:5}
\end{figure*}

\textbf{Our framework:} In this paper, we propose Optimization-Free Test-Time Adaptation for test-time adaptation on HAR. Before we introduce each component of the proposed OFTTA framework, we first give an overview of OFTTA, and the visualization is shown in Figure \ref{fig:5}. We are given an online mini-batch sensor data, the data goes through a feature extractor like CNNs and outputs high-dimensional features, where CBN has been replaced by our proposed EDTN. We first obtain pseudo labels via the former prototype-based classifier. The support set is updated by features based on their pseudo labels and filtered via entropy ranking. The adjusted prototype is computed by the centroid per class in the support set. Finally, we obtain the adjusted prediction.

\subsection{Feature Extractor Adjustment via Exponential Decay Test Batch Normalization}
For the feature extractor, we focus on adjusting the batch normalization layer in CNNs, which is commonly used to improve performance and stability. Specifically, considering a mini-batch feature $\mathbf{z}\in R^{B\times C\times H\times W}$, where $B$ is the batch size, $C$ is the number of channels, $H$ and $W$ are the height and width of feature input, we have the first and second order statistics of this batch:

\begin{equation}
\mu_c = \frac{1}{B \cdot H \cdot W} \sum_{b=1}^B \sum_{h=1}^H \sum_{w=1}^W \mathbf{z}_{bchw}, \qquad \sigma_c^2 = \frac{1}{B \cdot H \cdot W} \sum_{b=1}^B \sum_{h=1}^H \sum_{w=1}^W (\mathbf{z}_{bchw} - \mu_c)^2,
\label{eq:3}
\end{equation}
where $\mu_c$ and $\sigma_c^2$ are the mean and variance in the $c$-th channel of the input feature, respectively. The standardization  formula is as follows:
\begin{equation}
\hat{\mathbf{z}}_{bchw} = \frac{\mathbf{z}_{bchw} - \mu_c}{\sqrt{\sigma_c^2}},
\label{eq:4}
\end{equation}

Using BN layers can make neural network training faster and more stable. In the i.i.d. assumption, we use $\mu_s$ and $\sigma_s^2$ estimated from the source domain to standardize the target domain data.
However, it is not always the case when the input distribution changes. If the target domain data and the source domain data are not from the same distribution, the estimated statistics $\mu_s$ and $\sigma_s^2$ would be not reliable and the feature map can not be standardized correctly. To address distribution shifts in the testing phase, some popular methods \cite{schneider2020improving,you2021test, lim2023ttn} combine the conventional batch normalization (CBN) statistics ${{\mu }_{s}}$,${\sigma }_{s}^{2}$ with test batch normalization (TBN) statistics ${{\mu }_{t}}$, ${\sigma }_{t}^{2}$. The reestimated BN statistics are written as follows: 
\begin{equation}
\tilde{\mu }=\alpha {{\mu }_{s}}+(1-\alpha ){{\mu }_{t}}, \qquad {{\tilde{\sigma }}^{2}}=\alpha {\sigma }_{s}^{2}+(1-\alpha ){\sigma }_{t}^{2},
\label{eq:5}
\end{equation}
\noindent where $\alpha \in {{\mathbb{R}}^{C}}$ is a prior ratio that controls the statistic trade-off between CBN with TBN. The detailed algorithm description of EDTN is presented in Algorithm \ref{alg:1}.
Although previous methods have achieved considerable performance in TTA, there are still existing problems. (1) \textbf{Sensitive Hyperparameters}: In  \cite{schneider2020improving,you2021test}, the authors used the predefined ratio to mix the statistics. However, they do not consider the internal relationships between layers, and the best mix ratio may not be the same when the distribution shifts are different. In the HAR scenario, it is hard for us to adjust the ratio according to inter-subject variability. (2) \textbf{Privacy Concerns}: In TTN \cite{lim2023ttn}, the authors considered adding a post-training stage before the adaptation. Although the authors adjusted the model in a channel-wise manner, they use the source data and label in this stage. Due to the potential privacy issue in HAR, we can not access data from subjects in the source domain. (3) \textbf{Computational Cost}: In Mix-norm \cite{hu2021mixnorm}, the author obtained more accurate statistics via consistency between augmentation features. Nevertheless, in real-time HAR inference, we can not infer the model multiple times due to computation-restricted devices. Moreover, complex data augmentation techniques may not be applied to sensor data since sensor data is sensitive to temporal relations. Intuitively, it is natural to ask: \textbf{How to adjust the mix-ratio in sensor-based HAR problems?} We introduce a particularly straightforward layer-wise strategy to set the mix-ratio called Exponential Decay Test-time Normalization (EDTN).
Assuming there are $n$ BN layers, we have the prior ratio at layer $i$:
\begin{equation}
{\alpha}(i) = 
\begin{cases}
   \lambda^{\,i-n} &\text{if } i < n, \\
  1 &\text{if } i = n,
\end{cases}
\label{eq:6}
\end{equation}
where $\alpha(i)$ is the trade-off ratio between CBN and TBN at $i$ layer and $\lambda$ denotes the decay factor. In other words, TBN  dominates at the beginning, but as the layers go deeper, CBN takes over. 

\begin{figure}[t!]
	\vspace{-.2in}
	\centering
	\includegraphics[scale=0.5]{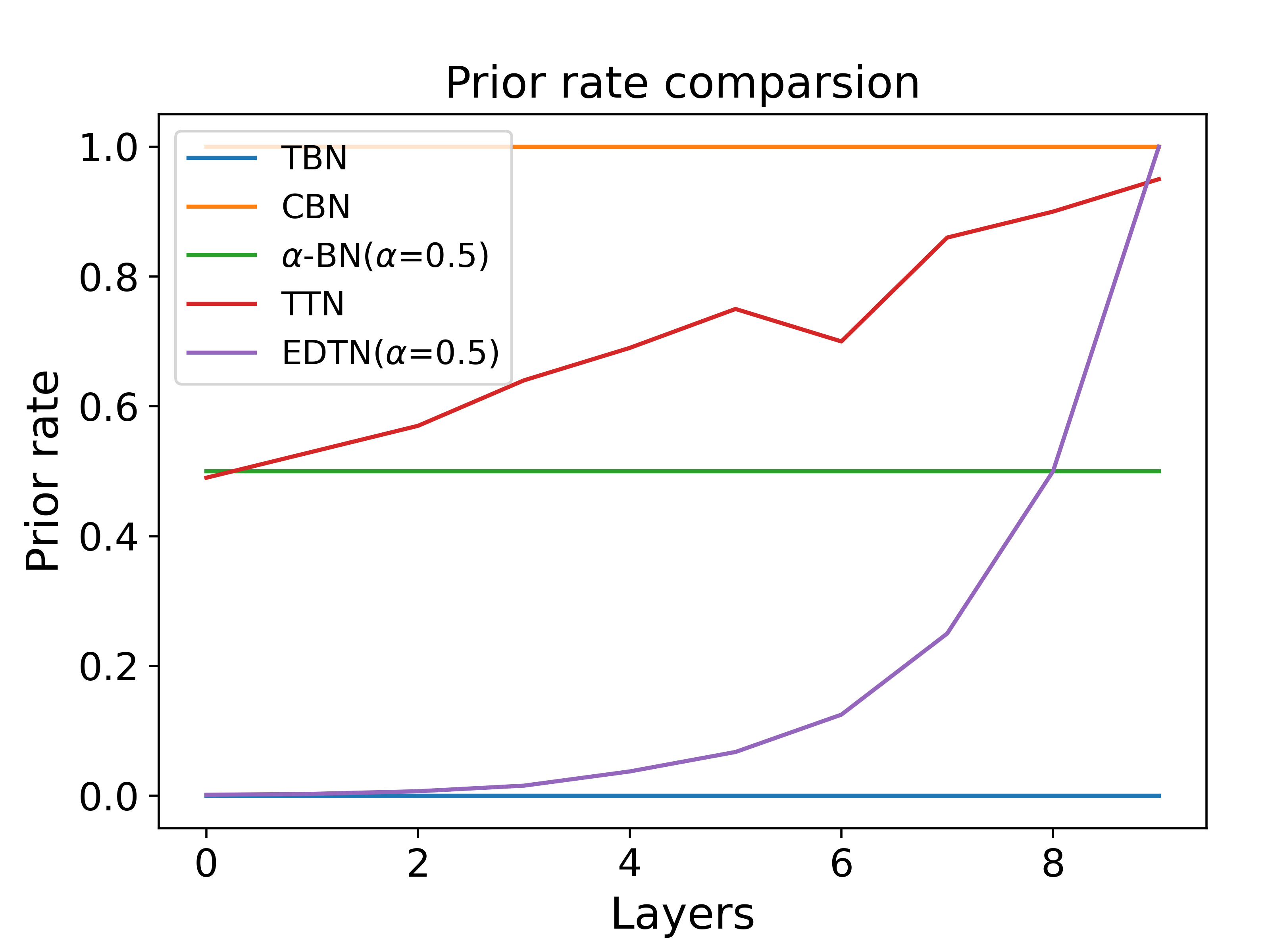}
	\caption{\label{Fig02} \normalsize{Prior rate comparison of a 10-BN-layer model. As the prior rate increases, the proportion of BN statistics from the source domain becomes higher. TTN is a score-based mix strategy. The rest of the methods are predefined hyperparameters. Once the adaptation starts, the ratio no longer changes.}}
\label{fig:6}
\end{figure}

\noindent \textbf{Remarks.} EDTN can solve or mitigate the problems discussed above: (1) For hyperparameter sensitivity, we just use a reasonable $\lambda$ (e.g., 0, 0.3, 0.7, etc.) and will improve the adaptation performance. For privacy exposure issues, EDTN does not need data from the source domain to set prior ratio. For computation cost, we just replace the pre-trained BN layer with our EDTN with no extra computational cost. For direct comparison, we visualized the difference between EDTN and other mix strategies in Figure \ref{fig:6}. 

\begin{algorithm}[t!]
\renewcommand{\algorithmicensure}{\textbf{Output:}}
\caption{\textbf{E}xponential \textbf{D}ecay \textbf{T}est-time \textbf{N}ormalization (EDTN) module \label{alg:1}}
\label{alg:batch_norm}
\begin{algorithmic}[1]
\REQUIRE Input feature $z \in \mathbb{R}^{B\times C\times H\times W}$, with batch size $B$, $C$ channels, height $H$ and width $W$; affine parameters $\gamma \in \mathbb{R}^C$, $\beta \in \mathbb{R}^C$; conventional batch normalization(CBN) mean ${{\mu }_{s}} \in \mathbb{R}^C$, variance ${\sigma }_{s}^{2} \in \mathbb{R}^C$   ; current layer prior $\alpha$

\STATE $ \mu_t = \frac{1}{B\times H \times W}\sum_{i=1}^B\sum_{j=1}^H\sum_{k=1}^W z_{i,:,j,k}$ // \textcolor{gray}{get test-time mean}   

\STATE $ \sigma_t^2 = \frac{1}{B\times H \times W}\sum_{i=1}^B\sum_{j=1}^H\sum_{k=1}^W (z_{i,:,j,k}-\mu_t)^2$ // \textcolor{gray}{get test-time variance} 

\STATE $\tilde{\mu }=\alpha {{\mu }_{s}}+(1-\alpha ){{\mu }_{t}}$ // \textcolor{gray}{get mixed mean $\tilde{\mu }$} 

\STATE ${{\tilde{\sigma }}^{2}}=\alpha {\sigma }_{s}^{2}+(1-\alpha ){\sigma }_{t}^{2}$ // \textcolor{gray}{get mixed variance ${{\tilde{\sigma }}^{2}}$}

\STATE ${{z}^{*}}=\frac{z-\tilde{\mu }}{\sqrt{{{{\tilde{\sigma }}}^{2}}}}$ // \textcolor{gray}{normalize}

\STATE ${{z}^{\star }}=\gamma \cdot {{z}^{*}}+\beta$ // \textcolor{gray}{scale and shift}
\ENSURE ${{z}^{\star}}$

\end{algorithmic}
\end{algorithm}

\subsection{Linear Classifier Adjustment via Prototype-based Classification}
For the linear classifier, we adjust prototype representations in the layer (the last layer of the model), which are computed by the centroid of each class in the high dimensional space. In the test time, we adjust the centroid using unlabeled test data and make real-time predictions. Specifically, inspired by T3A \cite{iwasawa2021test}, we also maintain a support set to adjust the centroid in an optimization-free online manner. We first introduce some necessary definitions. Consider a pre-trained model $f$ using feature extractor $h$ and an online new data ${x}_t$ at time $t$, we denote $h({x}_t)\in {{\mathbb{R}}^{m}}$ as the feature vector extracted by the penultimate layer of the model, which encodes ${x}_t$ to dimension $m$ in the feature space. One weight matrix $\mathbf{W}\in {{\mathbb{R}}^{m\times K}}$ in the classifier connects the feature $h(x)$ and the output $f(x)$ via linear operation $f({x}_t)={{\mathbf{W}}^{\text{T}}}{x}_t+\text{b}$, where $K$ is the total class number and $\text{b}\in {{\mathbb{R}}^{K}}$ is the bias vector. Since the support set is empty at the beginning of the adaptation, we initialize it as:
\begin{equation}
\mathbb{S}_{0}^{k}=\left\{ \frac{{{w}^{k}}}{\left\| {{w}^{k}} \right\|} \right\},
\label{eq:7}
\end{equation}
where ${{w}^{k}}\in \mathbf{W}\subset {{\mathbb{R}}^{m\times 1}}$ is the weight belonging to $k$-th class for $k=1,2,...K$ and $\left\| \cdot  \right\|$ is the L2 normalization. At time $t$, given the input ${x}_t$ we first get the feature ${f}({{x}_{t}})$ and its corresponding pseudo label:
\begin{equation}
\hat{y}=\arg {{\max }_{k}}{{f}_{k}}({{x}_{t}}),
\label{eq:8}
\end{equation}
where ${f}_{k}$ is the logits of the $k$-th class. The support set will be updated as:
\begin{equation}
\mathbb{S}_{t}^{k}=\left\{ \begin{array}{*{35}{l}}
   \mathbb{S}_{t-1}^{k}\cup \left\{ \frac{h({{x}_{t}})}{\left\| h({{x}_{t}}) \right\|} \right\} & &\text{if} \;\hat{y}=k  \\
   \mathbb{S}_{t-1}^{k} & &\text{otherwise}. 
\label{eq:9}
\end{array} \right. \end{equation}

We can obtain the updated class prototype for class $k$ using the centroid ${{\mu }_{k}}$, which can be computed as:
\begin{equation}
{{\mu }_{k}}=\frac{1}{|\mathbb{S}_{t}^{k}|}\sum\limits_{\mathbf{z}\in \;\mathbb{S}_{t}^{k}}{\mathbf{z}}.
\label{eq:10}
\end{equation}

Finally, we can compute the adjusted output by assigning features to the nearest prototypes:
\begin{equation}
{\hat{y}}'=\arg {{\min }_{k}}d(h({{x}_{t}}),{{\mu }_{k}}),
\label{eq:11}
\end{equation}

where $d(\cdot)$ is a distance metric. We use the cosine similarity as the distance metric $d(\cdot)$ throughout this paper. Since we use the pseudo label as the criterion to update the support set, it is inevitable that some instances will be assigned with incorrect labels and hinder the model from a more optimal centroid. To avoid this issue, we follow standard strategy \cite{iwasawa2021test, lim2023ttn}, which only uses partial support set via prediction entropy ranking: 
\begin{equation}
\mathbb{S}_{t}^{k}=\{z|z\in \mathbb{S}_{t}^{k},{{H}_{{{\text{W}}_{t}}}}(z)\le {{\beta }^{k}}\},
\label{eq:12}
\end{equation}
where ${\beta }^{k}$ is the $M$-th largest prediction entropy in ${S}_{t}^{k}$ and ${H}_{{{\text{W}}_{t}}}$ is the classifier-softmax-entropy function with parameter ${{{\text{W}}_{t}}}$ at time $t$. The whole procedure is described in Algorithm \ref{alg:2}.

\begin{figure}[htbp]
  \centering
  \begin{minipage}[t]{0.48\textwidth}
    \centering
    \includegraphics[width=\textwidth]{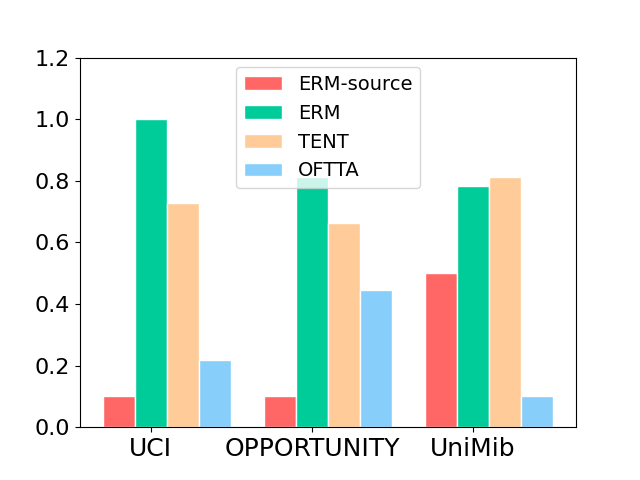}
    \caption{Prediction entropy comparison of ERM on source data, ERM on target data, TENT and OFTTA. OFTTA reduces prediction entropy in an optimization-free manner.}
    \label{fig:7}
  \end{minipage}
  \hfill
  \begin{minipage}[t]{0.48\textwidth}
    \centering
    \includegraphics[width=\textwidth]{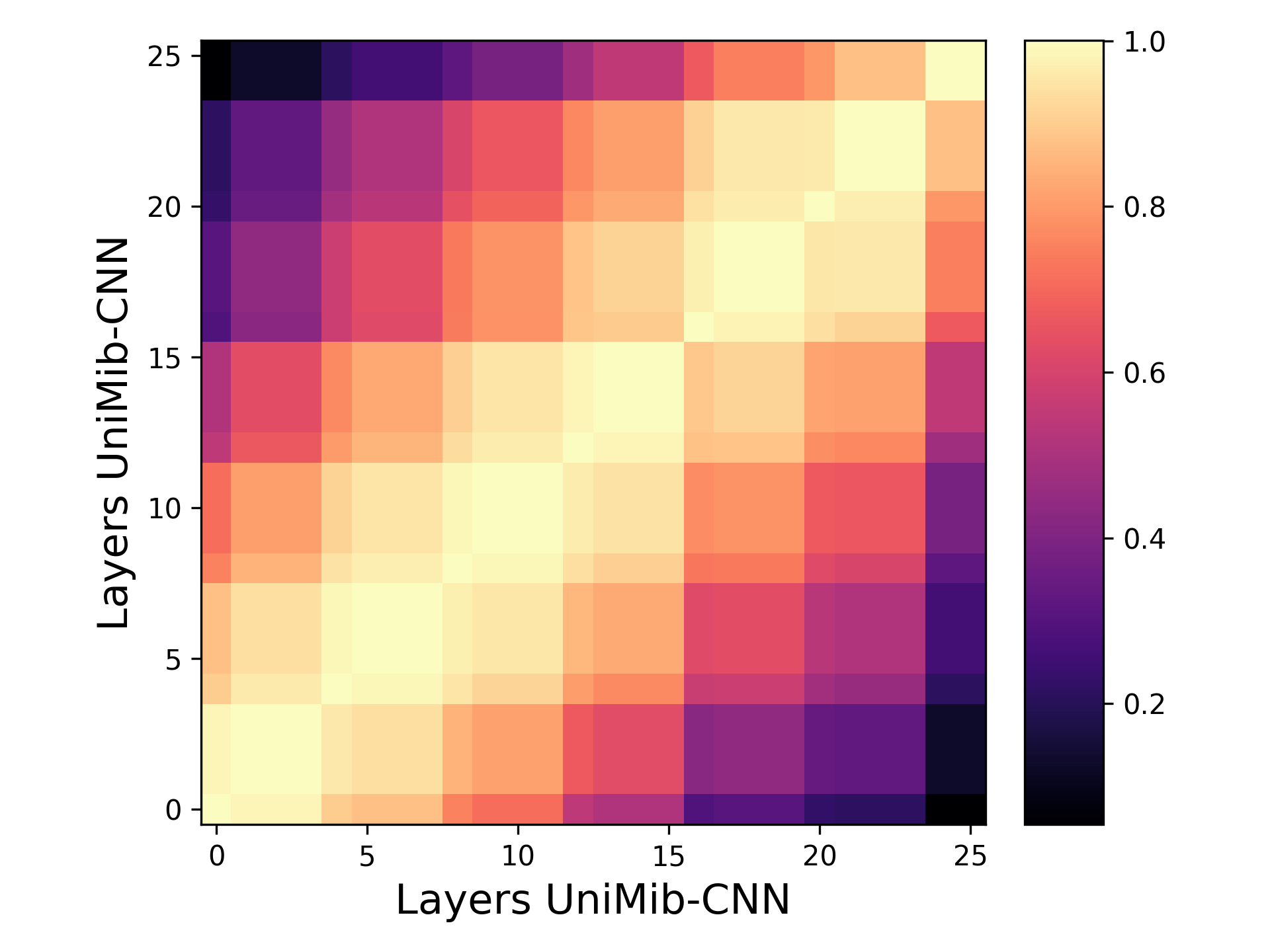}
    \caption{Layer-wise feature similarity visualization using CKA heatmap. UniMib-HAR dataset is used and the model is CNNs. x and y axes index the layers from input to output.  A higher score indicates greater similarity between the two features.}
    \label{fig:8}
  \end{minipage}
\end{figure}

\noindent\textbf{Remarks.} It can be easily observed that the prototype-based adaptation strategy has two advantages for sensor-based HAR. (1) \textbf{Implicit Entropy Minimization}: As the previous work suggests \cite{iwasawa2021test,niu2022efficient, wang2020tent}, if the model has low confidence on the input data, or in other words, the Shannon entropy \cite{lin1991divergence} of logits is high, the model tends to make incorrect predictions. To avoid this issue, TENT adjusts the model in an explicit manner: minimizing the prediction entropy via updating the affine parameters in normalization layers. However, adjusting parameters aggressively may cause the model to overfit the covariate noise in the adaptation flow. In contrast, our framework decreases the entropy of the logits over three HAR datasets in an implicit manner: using low-entropy features in the support set to calculate a more reliable centroid and further make more confident predictions. We visualize the prediction entropy in Figure \ref{fig:7}, the entropy of the data from an unseen domain tends to be higher than the source domain. Note that our OFTTA reduces the prediction entropy via implicit entropy minimization. (2) \textbf{Computational Cost}: As we mentioned in Figure \ref{fig:3}, we adjust the linear classifier in an optimization-free manner, which can greatly reduce memory and computational cost. Moreover, for each class, only $M$ features will be stored, and we do not need to worry about infinite memory growth since the length of the support set is fixed.

\begin{algorithm}[H]
\renewcommand{\algorithmicensure}{\textbf{Output:}}
\caption{Classifier Adjustment via Prototype-based Classification \label{alg:2}}

\begin{algorithmic}[1]
\REQUIRE Feature extractor $h$; test mini-batch ${\mathbb{B}}$; support set $\mathbb{S}_{t-1}^{k}$ at time $t$; number of samples per class in support set $M$
\STATE Obtain the mini-batch pseudo labels $\hat{y}$ via eq. \ref{eq:8}
\STATE Obtain updated support set $\mathbb{S}_{t}^{k}$ with eq. \ref{eq:9}
\STATE Filter the updated support set using eq. \ref{eq:12}
\STATE Compute the adjusted prototype with eq. \ref{eq:10}
\STATE Obtain the adjusted prediction ${\hat{y}}'$ with distance-based prototype with eq. \ref{eq:11}
\ENSURE ${\hat{y}}'$, $\mathbb{S}_{t}^{k}$
\end{algorithmic}
\end{algorithm}

\subsection{Discussions}
In this section, we mainly answer the motivation of the OFTTA framework and our insights into its components. Two recent works are also discussed in detail.

\textbf{(1) Why use TBN in a layer-wise decay manner?} Empirically, we use less TBN as the layer goes deeper. We can explain this strategy from the representation structure of modern neural networks \cite{raghu2021vision, paul2022vision}. We first visualize the feature map difference among layers from shallow to deep and plot Centered Kernel Alignment (CKA) similarity score \cite{kornblith2019similarity} in Figure \ref{fig:8}. We can observe that great diversity exists between lower and higher layers, which means features extracted from shallow and deep layers can be quite different. The phenomenon can be attributed to the limitation of kernel receptive fields. In shallow layers, the convolution layer can only access local information from the sensor data. As the layer goes deeper, the receptive field gradually expands to the entire sensor data segment and the convolutional layer can access global information, explaining the difference between shallow and deep features. In HAR, the diversity of personal habits such as body shapes, ages, etc., can lead to significant differences in activity details, making local features sensitive to domain shift. However, as the receptive field expands, deep layers can learn from a global perspective, which makes it easier to capture activity classes. Hence, the BN layer statistics in deep layers can be less sensitive to domain shift. Moreover, the layer-wise method can be regarded as a special form of the channel-wise method, but it is impractical to estimate the score for each channel without gradient descent \cite{lim2023ttn, niu2022efficient} since every channel needs to be treated equally from the perspective of the layer. In summary, we encourage decreasing the ratio of TBN in a layer-wise manner and using more CBN in the deep layers.

(2) \textbf{Why feature extractor and the classifier can be adjusted separately and work simultaneously?} In raw prototype adjustment \cite{iwasawa2021test}, the authors froze the feature extractor and only adjust the linear classifier. We empirically find that this method may not be not optimal. Since the support set can benefit from EDTN where the centroids of the prototypes, or rather the class weights of the non-parametric classifier, are calculated based on pseudo-labels. Through the technical lens, as prior works \cite{wang2020tent, niu2022efficient} suggest, the prediction entropy is often related to the error, as more confident predictions tend to be more correct. As is shown in Figure \ref{fig:7}, our proposed framework greatly reduces the prediction entropy since the low-entropy features in the support could be more accurately attributed to the correct prototypes. An empirical result is shown in Figure \ref{fig:10}. When equipped with EDTN, the classification results can noticeably improve by the non-parametric classifier and outperform other TTA methods.
Hence, we replace the pre-trained BN statistics with a combination of CBN and TBN to output more reliable features and further adjust the classifier. The results are discussed in detail in Section 5.1.

(3) \textbf{Complexity Analysis:} As we discussed above, our proposed framework adapts distribution shifts in an optimization-free manner. We first give an intuitive experiment in Figure \ref{fig:3} that gradient optimization can bring more memory cost and lead to slower inference speed. From the perspective of computational complexity, we incorporate TBN into the batch statistics to extract more robust features in the encoder. This operation only contains linear interpolation, so the increase in computational cost during inference is negligible. For the classifier, we only modify the linear classifier using a prototype-based support set. The extra computational overhead is the cost of one forward propagation of the last linear layer, which is tolerable compared with the cost of backpropagation. For the memory cost, we maintain the support set with a limited number of features. The resulting memory consumption is still lower than the cache generated by back propagation. The empirical results can be further illustrated in Section 6.

(4) \textbf{Relation to other prototype-based TTA methods}: Following T3A \cite{iwasawa2021test}, TAST \cite{jang2022test} also uses a support set to memorize features and construct the prototype. As we mentioned in Discussion (2), the prototype-based classification may overfit inaccurate pseudo labels and output an unreliable prototype. To tackle this issue, TAST  adds multi-trainable adaptation modules on top of the feature extractor and optimizes them via a distribution difference among modules' output. One key difference is that TAST freezes the feature extractor. In contrast, our OFTTA aims to mitigate this problem by adjusting the features before the classifier. By mixing CBN and TBN in the normalization layers, the extractor can output a more reliable feature and further benefit prototype adjustment. In essence, both methods share the motivation: mitigate overfitting problems caused by pseudo labels. However, TAST needs to train the adaptation modules via gradient descent and fine-tune multiple hyperparameters. Our method is performed in an optimization-free manner and uses quite simple hyperparameters, which separately adjust the feature extractor and classifier, which can easily benefit real-world HAR applications. We will further discuss this method in the experiments. 

(5) \textbf{Relation to other optimization-free TTA methods}: For efficient TTA, AdaNPC \cite{zhang2023adanpc} also proposed a quite simple method without parameter optimization. The linear classifier is simply replaced by a K-Nearest Neighbor (KNN) classifier. In the testing phase, the model obtains the prediction by voting among $k$ closed samples from the memory bank. Trustworthy features and predictions will be stored in the memory bank. It is obvious that AdaNPC shares two advantages with our OFTTA framework. \textbf{$(\text{I})$ Simple architecture:} Both methods adapt the model via a straightforward strategy with few insensitive hyperparameters. \textbf{$(\text{II})$ Efficient inference:} Both methods adapt models using a non-parametric paradigm, which can infer quickly in real-world applications. However, AdaNPC needs to memorize features and labels from the source domain, which are difficult to obtain in practical downstream tasks. In contrast, following fully test-time adaptation (FTTA) protocol \cite{wang2022continual, su2022revisiting, niu2023towards}, we focus on FTTA and only memorize samples from the target domain in this paper.

\section{EXPERIMENT}
In this section, we evaluate our OFTTA framework via extensive experiments on cross-person HAR dataset.

\subsection{Datasets and Preprocessing}
Following  \cite{qian2021latent}, we evaluate our method on three public sensor-based HAR datasets: UCI-HAR \cite{anguita2012human}, OPPORTUNITY \cite{roggen2010collecting}, and UniMiB-SHAR \cite{micucci2017unimib}. We list the main statistical information of the three datasets in Table \ref{tab:1}. The brief introduction for each dataset is as follows:
\begin{enumerate}
    \item \textbf{UCI-HAR (\textit{UCI})} \cite{anguita2012human}: The UCI dataset records 6 types pf daily activities(walking downstairs,  upstairs, sitting, standing, lying, walking) using accelerometer and gyroscope embedded in smart phone. 30 volunteers (19–48 years old) join the data collection phase with the phone attached on their waists. The smartphone captured 3-axial linear acceleration and 3-axial angular velocity at a constant rate of 50Hz.
    \item \textbf{OPPORTUNITY (\textit{OPPO})}  \cite{roggen2010collecting}: The Opportunity dataset was collected in a sensor-rich environment and consists of 15 wireless and wired networked sensor systems, with 72 sensors of 10 modalities sampled at 30 Hz. It includes recordings of accelerometer, gyroscope, magnetometer, and inertial measurement unit (IMU) data, and features 17 types of activities performed by four subjects in a breakfast preparation scenario. Opportunity dataset is highly detailed, with a total of 869,387 samples and a feature dimension of 113.
    \item \textbf{UniMiB-SHAR (\textit{UniMiB})}  \cite{micucci2017unimib}: The UniMiB SHAR dataset, conceived by Daniela et al., is a benchmark for human activity recognition and fall detection. The dataset includes data from 30 subjects aged 18 to 60 years, performing 17 fine-grained activities, which are divided into 9 types of activities of daily living (ADL) and 8 types of falls. Data collection was facilitated by an acceleration sensor embedded in an Android phone placed in the front trouser pocket of each participant. The sensor captured data at a sampling rate of 50Hz and each sample contained 3 vectors of 151 accelerometer values.
\end{enumerate}

\begin{table}[t!]
\caption{Statistical information of three HAR datasets.}
\vspace{-.1in}
\resizebox{.5\textwidth}{!}{
\begin{tabular}{ccccc}
\toprule
Dataset & Subject & Activity & Sample    & Sampling rate \\ \midrule
UCI     & 30      & 6        & 1,318,272 & 50 Hz         \\ 
OPPO    & 4       & 17       & 869,387   & 30 Hz         \\ 
UniMiB  & 30      & 17       & 11,771    & 50 Hz         \\ \bottomrule
\label{tab:1}
\end{tabular}
}
\vspace{-.1in}
\end{table}

\subsection{Experimental Setup}
\subsubsection{Dataset Pre-processing}
We construct our experiment under cross-person HAR setting. The detailed dataset pre-processing and the setting of cross-person domain HAR are summarized in Table \ref{tab:2}. To feed time-series data into the deep neural network, we first segment the sensor data by a standard sliding window technique \cite{bulling2014tutorial}. Since the window size and overlap may highly influence the classification performance, we utilized the previous standard setting to pre-process the sensor data \cite{qian2021latent}. The window size is set based on the dataset with 50\% overlap in adjacent windows.  All datasets have been normalized after segmentation. We follow the standard strategy to construct \cite{qian2021latent} the cross-person subset for each domain. For all three datasets, each domain only contains data from one individual subject. Specifically, for the UCI dataset, we choose the first 5 subjects' data (ID: 0, 1, 2, 3, and 4). For OPPO dataset, since only 4 subjects are involved in data collection, each subject's data can be treated as one domain. For the UniMiB dataset, 4 subjects' data (ID: 1, 2, 3, and 5) are used since those subjects conducted all 17 activities during the data collection.

\begin{table}[t!]
\caption{Detailed dataset information on cross-person HAR dataset.}
\vspace{-.1in}
\resizebox{.8\textwidth}{!}{
\begin{tabular}{ccccccccc}
\toprule
Dataset & Window size & Instance second (s) & Overlap rate & Domain & Domain ID        & Domain sample             & Total \\ \midrule
UCI     & 128 × 9     & 2.56  & 50\%         & 5      & (0; 1; 2; 3; 4)  & (347; 302; 341; 317; 302) & 1299  \\ 
OPPO    & 30 × 77     & 1.00  & 50\%         & 4      & (S1; S2; S3; S4) & (4344; 4206; 4125; 3417)  & 16092 \\ 
UniMiB  & 151 × 3     & 3.02  & 50\%         & 4      & (1; 2; 3; 5)     & (384; 583; 304; 298)      & 1569  \\ \bottomrule
\label{tab:2}
\end{tabular}
}
\vspace{-.1in}
\end{table}

\subsubsection{Model Architecture}
We mainly use the shallow CNNs as the main backbone in our experiment. Our algorithm is agnostic to specific network architectures. Shallow CNNs are computing-friendly and are widely used in recent sensor-based HAR research \cite{lu2022semantic, qian2021latent, huang2022deep, huang2022channel}. For efficiency and simplicity, specifically, we adopt 3-layer CNNs with \textit{Conv-BN-ReLU} module. The detailed information of the model is illustrated in Table \ref{tab:3}. We further discuss the architecture and replace the CNNs with other backbones in section \ref{sec:5.4}.

\begin{table}[t!]
\centering
\caption{Information on the architectures of the models.}

\vspace{-.1in}
\resizebox{.6\textwidth}{!}{
\begin{tabular}{cccccc}
\toprule
Dataset & Input        & Kernel size & Layer-1 & Layer-2 & Layer-3 \\ \midrule
UCI     & (1, 128, 9)  & (6, 1)      & $C$(64)   & $C$(128)  & $C$(256)  \\ 
OPPO    & (1, 30, 77)  & (9, 5)      & $C$(64)   & $C$(128)  & $C$(256)  \\ 
UniMiB  & (1, 151, 3) & (6, 1)      & $C$(128)  & $C$(256)  & $C$(384)  \\ \bottomrule
\label{tab:3}
\end{tabular}
}
\end{table}

\subsubsection{Comparison Methods}
We compared our OFTTA framework with several state-of-the-art TTA methods. A brief introduction of each method is shown below: 
\begin{enumerate}
    \item \textbf{Empirical Risk Minimization} (\textbf{ERM} \cite{vapnik1991principles}, i.e., CNN baseline):  minimizes classification error over source domains. ERM freezes the model in the testing phase. We regard it as the baseline.
    \item \textbf{BN} \cite{schneider2020improving}: incorporates test-time batch statistics into pre-trained BN layer. 
    \item \textbf{T3A} \cite{iwasawa2021test}: adjusts the classifier by computing class prototypes.
    \item \textbf{TENT} \cite{wang2020tent}: minimizes the test mini-batch entropy via unsupervised entropy loss. Only affine parameters in BN layers will be updated.
    \item \textbf{PL} \cite{lee2013pseudo}: retrains the model using online confident pseudo labels of the mini-batch.
    \item \textbf{SHOT} \cite{liang2020we}: adapts the model via exploiting entropy minimization, class-balanced regulation, and self-supervised pseudo-labeling.
    \item \textbf{SAR} \cite{niu2023towards}: combines entropy and sharp-aware optimization to encourage model in smooth loss landscape.
    \item \textbf{TAST} \cite{jang2022test}: utilizes trainable BatchEnsemble module to align the output distribution between a prototype-based classifier and a neighbor-based classifier.
    \item \textbf{TAST-BN} \cite{jang2022test}: is a variant of TAST and stores the test data itself in a support set instead of features.
\end{enumerate}

\subsubsection{Implementation Details}
(1) \textbf{Training phase}: Before we conduct the adaptation on the target domain, we need to pre-train a model on one or several source domains. To achieve stable performance on the source domain, we set the learning rate to $1{{e}^{-4}}$ and it decays by 50\% at every 20 epochs. The number of epochs is set to 100.
We set the batch size to 128, 512, and 64 for UCI, OPPO, and UniMiB, respectively.
We train the model by the rule of  ERM. Under the setting of the multi-source domain, the learning objective of ERM can be defined as minimizing the weighted average of losses over all source domains.
We employ the standard cross-entropy loss to optimize the model and save the checkpoint with the best validation loss for further adaptation.
We empirically verified that the model can converge and has stable performance on the source domain using these hyperparameters.
(2) \textbf{Testing phase}: For ERM, it still utilizes BN parameters trained on the source domains to normalize the features. For other TTA methods,  we followed the recent standard TTA setting and adapted the model using an online mini-batch in one-pass \cite{su2022revisiting, qin2022domain, niu2023towards}.
For BN, T3A, TENT, PL, and SHOT, we used the public code\footnote{\url{https://github.com/matsuolab/T3A}}. 
For SAR, we modified from their official implementation\footnote{\url{https://github.com/mr-eggplant/SAR}}.
Following their default setting, the reliable entropy ratio is set to 0.4.
For TAST and TAST-BN, we modified the official implementation\footnote{\url{https://github.com/mingukjang/TAST}}.
Since TAST has a lot of hyperparameters, we chosen the hyperparameters via grid search to find the best hyperparameter combination.
Empirically, the number of nearby support examples is set to 8 and the ensemble number is set to 10.
Moreover, all methods can be separated into two groups: Optimization-free (ERM, BN, T3A, OFTTA) and Optimization with gradient descent (TENT, PL, SHOT, SAR, TAST, TAST-BN).
For all methods with optimization, empirically, we used Adam optimizer with the learning rate of $1{{e}^{-2}}$.
For all BN-optimization methods (BN, TENT, PL, SHOT, SAR), following TENT\footnote{\url{https://github.com/DequanWang/tent}}, we only used TBN to normalize the features and affine parameters will be updated.
For all prototype-based methods (T3A, TAST, TAST-BN, OFTTA), we set the support examples to 25 and -1 in LOOA and CTTA for each class.
For our proposed OFTTA, we did not specify the decay factor $\lambda$.
As an alternative, we just set the prior ratio ${\alpha}(i)$ to 0.1 at the bottom layer and 1.0 at the top layer.
The decay factor depends on the layer number. The batch size of UCI, OPPO, UniMiB are set to 180, 220, and 1000, respectively.
Note that all methods were compared fairly under the same setting. We used PyTorch as the deep learning framework to implement all experiments.
The experiments were conducted on a server with a GPU of GeForce 3090Ti.

\begin{figure}[htbp]
  \centering
  \subfigure[Leave-One-Out Adaptation (LOOA)]{\includegraphics[width=0.35\linewidth]{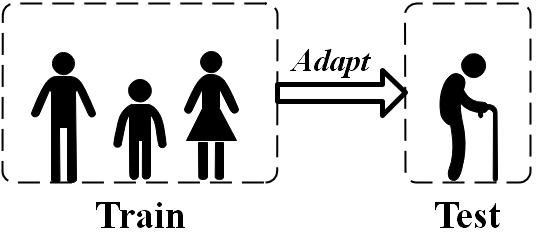}\label{fig:subfig1}}
  \hspace{0.1\linewidth}
  \subfigure[Continual Test-Time Adaptation (CTTA)]{\includegraphics[width=0.45\linewidth]{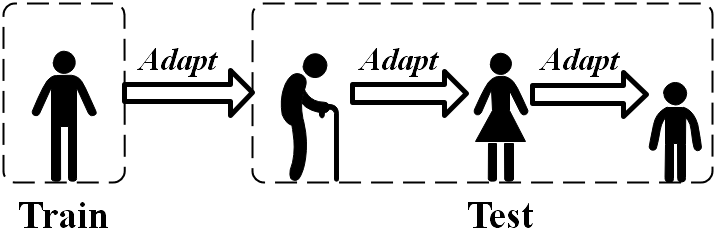}\label{fig:subfig2}}
  \caption{Visual description of LOOA and CTTA setting}
  \label{fig:9}
\end{figure}

\begin{table*}[t!]
\caption{Classification accuracy (\%) for leave-one-out adaptation. The \textbf{bold} and {\ul underline} items denote the best of all methods and the best results of optimization-free methods, respectively.}
\resizebox{1.0\textwidth}{!}{
\begin{tabular}{cc|cccccccccc}
\toprule
 & Tar & $\text{ERM}^\dagger$ & $\text{BN}^\dagger$ & $\text{T3A}^\dagger$ & $\text{TENT}^\psi$ & $\text{PL}^\psi$ & $\text{SHOT}^\psi$ & $\text{SAR}^\psi$ & $\text{TAST}^\psi$ & $\text{TAST-BN}^\psi$ & $\text{OFTTA}^\dagger$ \\
\midrule
\multirow{6}{*}{\rotatebox{90}{UCI}}
& 0 & 93.15 & 92.89 & 94.62 (0.17) & 94.23 (0.29) & 93.85 (0.17) & 94.14 (0.17) & 92.51 (0.86) & 94.72 (0.17) & \textbf{95.29 (0.15)} & \underline{95.20 (0.17)} \\
& 1 & 85.10 & 84.33 & 87.75 (0.57) & 84.77 (0.58) & 83.99 (2.49) & 83.99 (2.78) & 84.22 (2.10) & \textbf{88.19 (0.50)} & 87.39 (0.32) & \underline{87.64 (0.19)} \\
& 2 & 97.54 & 98.93 & 99.02 (0.34) & 99.12 (0.29) & 98.63 (0.17) & 98.73 (0.17) & 98.53 (0.51) & \textbf{99.32 (0.17)} & 99.28 (0.12) & \underline{98.83 (0.51)} \\
& 3 & 78.80 & 83.07 & 78.65 (2.10) & 82.86 (1.93) & 82.86 (1.01) & 83.07 (0.91) & 82.44 (1.31) & 80.02 (1.19) & \textbf{83.42 (2.08)} & \underline{83.49 (1.74)} \\
& 4 & 76.82 & 85.98 & 86.75 (0.33) & 88.74 (0.66) & 86.31 (0.50) & 86.43 (0.57) & 85.43 (0.99) & 86.75 (0.57) & 90.31 (1.81) & \underline{\textbf{91.17 (1.64)}} \\
& AVG & 86.28 & 89.04 & 89.36 (0.53) & 89.94 (0.57) & 89.13 (0.43) & 89.27 (0.47) & 88.62 (0.29) & 89.80 (0.38) & 91.14 (0.87) & \underline{\textbf{91.26 (0.78)}} \\
\midrule
\multirow{5}{*}{\rotatebox{90}{OPPO}} 
& S1 & 83.54 & 83.77 & \underline{82.87 (0.58)} & 83.44 (0.23) & 83.58 (0.09) & \textbf{84.06 (0.11)} & 83.71 (0.09) & 83.93 (0.67) & 83.94 (0.50) & 82.83 (0.54) \\
& S2 & 73.25 & 73.99 & \underline{74.91 (0.28)} & 74.75 (0.44) & 74.90 (0.21) & 74.86 (0.12) & 73.87 (0.08) & \textbf{76.17 (0.33)} & 75.85 (0.25) & 74.64 (0.65) \\
& S3 & 70.21 & 70.59 & 71.72 (0.67) & 70.08 (0.23) & 70.79 (0.10) & 71.17 (0.27) & 70.63 (0.03) & \textbf{73.03 (0.23)} & 72.70 (0.12) & \underline{71.89 (0.55)} \\
& S4 & 80.33 & 81.05 & 80.84 (0.41) & 81.17 (0.35) & 81.27 (0.11) & 81.10 (0.15) & 81.07 (0.12) & 81.13 (0.54) & \textbf{81.74 (0.84)} & \underline{81.24 (0.30)} \\
& AVG & 76.83 & 77.35 & 77.59 (0.15) & 77.36 (0.05) & 77.64 (0.03) & 77.80 (0.03) & 77.32 (0.02) & \textbf{78.57 (0.28)} & 78.56 (0.39) & \underline{77.65 (0.23)} \\
\midrule
\multirow{5}{*}{\rotatebox{90}{UniMiB}} 
& 1 & 60.42 & 65.97 & 61.72 (0.26) & 64.93 (0.66) & 66.06 (0.15) & 66.14 (0.26) & 66.06 (0.30) & 62.93 (0.30) & 68.06 (0.14) & \underline{\textbf{68.11 (1.07)}} \\
& 2 & 66.39 & 69.64 & 66.84 (2.55) & 70.16 (0.95) & 69.70 (1.14) & 69.87 (1.48) & 69.64 (1.19) & 69.41 (5.42) & 73.12 (1.43) & \underline{\textbf{73.76 (1.93)}} \\
& 3 & 75.66 & 75.33 & 80.48 (0.50) & 75.88 (0.68) & 75.22 (1.55) & 75.33 (1.18) & 75.33 (1.43) & 79.49 (0.76) & 77.14 (0.72) & \underline{\textbf{80.26 (0.57)}} \\
& 5 & 41.63 & 41.83 & 41.83 (0.39) & 42.28 (1.47) & 41.83 (1.08) & 41.83 (1.08) & 41.28 (1.46) & 41.95 (0.58) & 42.00 (0.67) & \underline{\textbf{45.30 (0.34)}} \\
& AVG & 61.02 & 63.19 & 62.72 (0.65) & 63.31 (0.31) & 63.20 (0.17) & 63.29 (0.30) & 63.08 (0.23) & 63.45 (1.57) & 65.08 (0.64) & \underline{\textbf{66.86 (0.33)}} \\
\midrule
& AVG all & 74.71 & 76.53 & 76.55 (0.16) & 76.87 (0.26) & 76.66 (0.17) & 76.79 (0.26) & 76.34 (0.07) & 77.27 (0.31) & 78.26 (0.25) & \underline{\textbf{78.59 (0.29)}} \\
\bottomrule
\multicolumn{8}{l}{$\dagger$: Optimization-free, $\psi$: Optimization with gradient descent.}\\
\end{tabular}
\label{tab:4}
}
\end{table*}

\begin{table*}[t!]
\caption{Macro-F1 score (\%) for leave-one-out adaptation. The \textbf{bold} and {\ul underline} items denote the best of all methods and the best results of optimization-free methods, respectively.}
\resizebox{1.0\textwidth}{!}{
\begin{tabular}{cc|cccccccccc}
\toprule
 & Tar & $\text{ERM}^\dagger$ & $\text{BN}^\dagger$ & $\text{T3A}^\dagger$ & $\text{TENT}^\psi$ & $\text{PL}^\psi$ & $\text{SHOT}^\psi$ & $\text{SAR}^\psi$ & $\text{TAST}^\psi$ & $\text{TAST-BN}^\psi$ & $\text{OFTTA}^\dagger$ \\
\midrule
\multirow{6}{*}{\rotatebox{90}{UCI}} 
& 0 & 92.18 & 92.02 & 94.00 (0.16) & 93.39 (0.34) & 93.00 (0.22) & 93.33 (0.18) & 91.64 (1.03) & 94.10 (0.15) & 94.33 (0.93) & \underline{\textbf{94.56 (0.19)}} \\
& 1 & 82.33 & 81.40 & \underline{85.38 (0.51)} & 81.71 (0.61) & 81.16 (2.24) & 81.19 (2.47) & 81.29 (1.95) & \textbf{86.06 (0.15)} & 85.18 (0.14) & 84.96 (0.28) \\
& 2 & 97.38 & 98.94 &\underline{99.04 (0.32)} & 99.11 (0.30) & 98.64 (0.18) & 98.74 (0.17) & 98.55 (0.49) & 99.31 (0.16) & \textbf{99.31 (0.09)} & 98.84 (0.50) \\
& 3 & 79.37 & 82.94 & 78.58 (1.94) & 82.77 (1.99) & 82.75 (1.00) & 82.96 (0.88) & 82.30 (1.32) & 79.91 (1.13) & 82.79 (2.43) & \underline{\textbf{83.44 (1.73)}} \\
& 4 & 76.03 & 86.01 & 86.61 (0.35) & 88.69 (0.66) & 86.33 (0.50) & 86.44 (0.57) & 85.47 (0.98) & 86.62 (0.60) & 89.79 (2.09) & \underline{\textbf{91.08 (1.59)}} \\
& AVG & 85.46 & 88.26 & 88.72 (0.47) & 89.14 (0.59) & 88.38 (0.38) & 88.53 (0.41) & 87.85 (0.34) & 89.20 (0.31) & 90.28 (1.03) & \underline{\textbf{90.58 (0.76)}} \\
\midrule
\multirow{5}{*}{\rotatebox{90}{OPPO}} 
& S1 & 77.25 & 77.50 & \underline{76.19 (0.74)} & 77.00 (0.30) & 77.18 (0.19) & 77.71 (0.15) & 77.38 (0.07) & 77.79 (0.80) & \textbf{77.91 (0.78)} & 76.18 (0.72) \\
& S2 & 66.78 & 69.00 & \underline{70.68 (0.50)} & 69.89 (0.56) & 70.03 (0.29) & 69.82 (0.18) & 68.79 (0.05) & \textbf{71.80 (0.53)} & 71.62 (0.47) & 70.31 (1.00) \\
& S3 & 59.33 & 60.24 & 64.70 (0.71) & 58.92 (0.20) & 60.22 (0.15) & 61.39 (0.17) & 60.28 (0.05) & \textbf{66.04 (0.24)} & 65.66 (0.40) & \underline{64.89 (0.56)} \\
& S4 & 75.63 & 76.54 & 77.09 (0.45) & 76.58 (0.44) & 76.70 (0.08) & 76.59 (0.21) & 76.58 (0.17) & 77.41 (0.72) & \textbf{77.66 (0.46)} & \underline{77.39 (0.50)} \\
& AVG & 69.75 & 70.82 & 72.17 (0.22) & 70.60 (0.05) & 71.03 (0.08) & 71.38 (0.04) & 70.76 (0.01) & \textbf{73.26 (0.42)} & 73.21 (0.33) & \underline{72.19 (0.40)} \\
\midrule
\multirow{5}{*}{\rotatebox{90}{UniMiB}} 
& 1 & 47.57 & 51.44 & 49.72 (0.32) & 49.86 (1.39) & 51.51 (2.04) & 51.53 (1.79) & 51.50 (1.89) & 48.20 (0.87) & 52.31 (0.44) & \underline{\textbf{54.45 (0.72)}} \\
& 2 & 57.22 & 58.90 & 63.07 (1.32) & 59.74 (0.46) & 58.99 (0.59) & 59.35 (0.76) & 58.91 (0.59) & 60.26 (2.02) & 66.44 (1.51) & \underline{\textbf{66.96 (1.24)}} \\
& 3 & 56.79 & 55.60 & 62.64 (1.37) & 56.03 (1.92) & 55.53 (2.96) & 55.61 (2.49) & 55.60 (2.85) & 58.11 (1.66) & 58.43 (0.68) & \underline{\textbf{61.78 (1.65)}} \\
& 5 & 28.53 & 24.93 & \underline{29.37 (0.26)} & 24.79 (0.43) & 24.92 (0.27) & 25.07 (0.22) & 24.43 (0.22) & \textbf{29.10 (1.03)}  & 28.44 (0.68) & 28.30 (0.05) \\
& AVG & 47.53 & 47.72 & 51.20 (0.44) & 47.61 (0.30) & 47.74 (0.26) & 47.89 (0.08) & 47.61 (0.07) & 48.92 (0.82) & 51.41 (0.63) & \underline{\textbf{52.87 (0.60)}} \\
\midrule
& AVG all & 67.58 & 68.93 & 70.70 (0.37) & 69.11 (0.31) & 69.05 (0.05) & 69.27 (0.12) & 68.74 (0.14) & 70.46 (0.30) & 71.63 (0.64) & \underline{\textbf{71.88 (0.58)}} \\
\bottomrule
\multicolumn{8}{l}{$\dagger$: Optimization-free, $\psi$: Optimization with gradient descent.}\\
\end{tabular}
\label{tab:5}
}
\end{table*}

\begin{table*}[t!]
\caption{Classification accuracy (\%) for continual test-time adaptation. The \textbf{bold} and {\ul underline} items denote the best of all methods and the best results of optimization-free methods, respectively.}
\resizebox{1.0\textwidth}{!}{
\begin{tabular}{cc|cccccccccc}
\toprule
 & Tar & $\text{ERM}^\dagger$ & $\text{BN}^\dagger$ & $\text{T3A}^\dagger$ & $\text{TENT}^\psi$ & $\text{PL}^\psi$ & $\text{SHOT}^\psi$ & $\text{SAR}^\psi$ & $\text{TAST}^\psi$ & $\text{TAST-BN}^\psi$ & $\text{OFTTA}^\dagger$ \\
\midrule
\multirow{6}{*}{\rotatebox{90}{UCI}} 
& 1,2,3,4  & 63.89 & 68.81 (0.00) & 65.99 (0.28) & 70.00 (1.34) & 69.65 (0.50) & 69.09 (0.62) & 68.69 (0.28) & 67.02 (0.06) & 69.37 (0.67) & \underline{\textbf{71.35 (0.23)}} \\
& 0,2,3,4  & 74.08 & \underline{77.82 (0.17)} & 74.36 (0.11) & 76.55 (0.95) & 77.97 (0.28) & \textbf{78.10 (0.78)} & 77.82 (0.29) & 74.13 (0.67) & 75.47 (0.34) & 75.71 (0.57) \\
& 0,1,3,4  & 65.25 & 67.74 (0.29) & 66.51 (0.23) & 67.42 (0.28) & 67.02 (0.28) & \textbf{68.53 (0.51)} & 67.94 (0.23) & 65.08 (0.33) & 68.49 (0.11) & \underline{68.30 (0.04)} \\
& 0,1,2,4  & 68.25 & 67.50 (0.17) & 72.06 (0.23) & 70.52 (0.28) & 69.88 (0.17) & 71.17 (0.08) & 67.34 (0.06) & 70.52 (0.39) & \textbf{72.58 (0.28)} & \underline{72.46 (0.45)} \\
& 0,1,2,3  & 57.74 & 68.42 (0.45) & 65.28 (0.28) & 70.46 (0.47) & 70.32 (0.90) & \textbf{73.46 (0.50)} & 68.38 (0.62) & 68.45 (1.74) & 70.83 (0.74) & \underline{70.28 (0.06)} \\
& AVG      & 65.82 & 70.06 (0.04) & 68.84 (0.11) & 70.99 (0.13) & 70.97 (0.13) & \textbf{72.07 (0.02)} & 70.03 (0.06) & 69.04 (0.35) & 71.35 (0.11) & \underline{71.62 (0.23)} \\
\midrule
\multirow{5}{*}{\rotatebox{90}{OPPO}} 
& S2,S3,S4 & 67.60 & \underline{67.87 (0.10)} & 67.29 (0.04) & \textbf{68.53 (0.18)} & 68.19 (0.23) & 67.69 (0.09) & 67.84 (0.01) & 63.89 (0.07) & 63.67 (0.80) & 67.38 (0.18) \\
& S1,S3,S4 & 65.01 & 65.36 (0.10) & \textbf{67.25 (0.25)} & 65.14 (0.45) & 65.34 (0.30) & 65.52 (0.17) & 65.32 (0.07) & 63.85 (0.52) & 64.09 (0.26) & \underline{67.10 (0.26)} \\
& S1,S2,S4 & 62.58 & 62.09 (0.04) & \underline{62.62 (0.38)} & 61.00 (0.36) & 61.46 (0.26) & \textbf{63.04 (0.11)} & 62.07 (0.03) & 58.07 (0.02) & 58.04 (0.08) & 62.38 (0.63) \\
& S1,S2,S3 & 67.55 & 68.42 (0.18) & 68.71 (0.16) & 67.91 (0.10) & 68.45 (0.04) & 68.08 (0.13) & 68.33 (0.14) & 64.36 (0.32) & 65.81 (2.08) & \underline{\textbf{68.81 (0.11)}} \\
& AVG      & 65.69 & 65.93 (0.05) & \underline{\textbf{66.46 (0.08)}} & 65.64 (0.05) & 65.86 (0.06) & 66.08 (0.04) & 65.89 (0.03) & 62.54 (0.07) & 62.90 (0.41) & 66.41 (0.16) \\
\midrule
\multirow{5}{*}{\rotatebox{90}{UniMiB}} 
& 2,3,5    & 31.91 & 32.59 (0.06) & 34.59 (0.83) & 31.37 (0.39) & 31.28 (0.52) & 32.64 (0.00) & 32.55 (0.00) & 31.96 (1.48) & \textbf{35.82 (0.26)} & \underline{35.54 (0.12)} \\
& 1,3,5    & 50.57 & 51.99 (0.40) & 46.02 (1.44) & 51.20 (0.56) & 51.93 (0.33) & 52.16 (0.16) & 51.99 (0.40) & 46.76 (1.05) & 55.46 (1.61) & \underline{\textbf{57.04 (0.43)}} \\
& 1,2,5    & 40.22 & 35.55 (0.01) & 39.32 (0.57) & 34.91 (0.25) & 35.96 (0.06) & 34.87 (0.19) & 35.59 (0.06) & \textbf{39.05 (0.57)} & 37.64 (0.39) & \underline{38.09 (0.16)} \\
& 1,2,3    & 26.56 & 24.82 (0.13) & \underline{27.41 (0.45)} & 21.41 (0.71) & 25.59 (0.33) & 24.23 (0.45) & 24.95 (0.07) & \textbf{28.82 (0.26)} & 25.73 (1.03) & 25.63 (0.08) \\
& AVG      & 37.32 & 36.23 (0.08) & 36.83 (0.18) & 34.72 (0.07) & 36.19 (0.14) & 35.97 (0.12) & 36.27 (0.10) & 36.64 (0.10) & 38.66 (0.18) & \underline{\textbf{39.08 (0.10)}} \\
\midrule
& AVG all & 56.27 & 57.41 (0.03) & 57.38 (0.05) & 57.12 (0.05) & 57.67 (0.11) & 58.04 (0.03) & 57.40 (0.06) & 56.07 (0.13) & 57.63 (0.11) & \underline{\textbf{59.04 (0.07)}} \\
\bottomrule
\multicolumn{8}{l}{$\dagger$: Optimization-free, $\psi$: Optimization with gradient descent.}\\
\end{tabular}
\label{tab:6}
}
\end{table*}

\begin{table*}[t!]
\caption{Macro-F1 score (\%) for continual test-time adaptation. The \textbf{bold} and {\ul underline} items denote the best of all methods and the best results of optimization-free methods, respectively.}
\resizebox{1.0\textwidth}{!}{
\begin{tabular}{cc|cccccccccc}
\toprule
 & Tar & $\text{ERM}^\dagger$ & $\text{BN}^\dagger$ & $\text{T3A}^\dagger$ & $\text{TENT}^\psi$ & $\text{PL}^\psi$ & $\text{SHOT}^\psi$ & $\text{SAR}^\psi$ & $\text{TAST}^\psi$ & $\text{TAST-BN}^\psi$ & $\text{OFTTA}^\dagger$ \\
\midrule
\multirow{6}{*}{\rotatebox{90}{UCI}} 
& 1,2,3,4  & 62.35 & 68.40 & 65.54 (0.21) & 69.34 (0.59) & 69.23 (0.13) & 68.96 (0.52) & 68.32 (0.23) & 66.71 (0.05) & 69.31 (0.65) &  \underline{\textbf{71.30 (0.24)}} \\
& 0,2,3,4  & 72.36 & 76.90 & 73.46 (0.16) & 75.64 (1.15) & 77.31 (0.23) & \textbf{77.45 (0.81)} & 77.05 (0.33) & 73.17 (0.66) & 75.24 (0.34) & \underline{75.44 (0.60)} \\
& 0,1,3,4  & 65.16 & 67.78 & 66.50 (0.26) & 67.57 (0.37) & 67.15 (0.34) & \textbf{68.87 (0.54)} & 68.20 (0.23) & 65.18 (0.21) & \underline{68.58 (0.14)} & 68.08 (0.41) \\
& 0,1,2,4  & 67.01 & 66.02 & 71.46 (0.28) & 68.77 (0.32) & 68.22 (0.28) & 69.68 (0.00) & 65.99 (0.04) & 69.98 (0.33) & \textbf{72.01 (0.34)} & \underline{71.81 (0.58)} \\
& 0,1,2,3  & 55.47 & 68.76 & 64.37 (0.37) & 70.19 (0.73) & 70.18 (0.95) & \textbf{73.62 (0.62)} & 68.33 (0.71) & 67.00 (1.29) & \underline{71.05 (0.91)} & 70.21 (0.04) \\
& AVG      & 64.47 & 69.57 & 68.26 (0.11) & 70.30 (0.40) & 70.42 (0.05) & \textbf{71.71 (0.04)} & 69.57 (0.07) & 68.40 (0.27) & 71.24 (0.16) & \underline{71.37 (0.21)} \\
\midrule
\multirow{5}{*}{\rotatebox{90}{OPPO}} 
& S2,S3,S4 & 61.63 & 62.01 & 61.64 (0.11) & \textbf{62.74 (0.26)} & 62.01 (0.36) & 62.06 (0.16) & 61.83 (0.12) & 57.42 (0.28) & 58.03 (0.11) & \underline{61.74 (0.27)} \\
& S1,S3,S4 & 55.95 & 56.18 & 58.79 (0.33) & 55.12 (0.55) & 55.80 (0.35) & 57.58 (0.24) & 56.32 (0.18) & 54.90 (0.65) & 55.06 (0.03) & \underline{\textbf{58.53 (0.42)}} \\
& S1,S2,S4 & 52.29 & 51.85 & 53.59 (0.47) & 49.86 (0.47) & 50.49 (0.02) & 53.74 (0.04) & 51.88 (0.10) & 48.12 (0.16) & 48.11 (0.13) & \underline{\textbf{54.08 (0.75)}} \\
& S1,S2,S3 & 58.47 & 59.75 & 59.66 (0.14) & 58.34 (0.08) & 59.45 (0.08) & 60.25 (0.06) & 59.57 (0.09) & 54.68 (0.40) & 54.91 (0.41) & \underline{\textbf{60.34 (0.13)}} \\
& AVG      & 57.09 & 57.45 & 58.42 (0.10) & 56.51 (0.02) & 56.93 (0.03) & 58.41 (0.02) & 57.40 (0.02) & 53.78 (0.17) & 54.03 (0.05) & \underline{\textbf{58.67 (0.18)}} \\
\midrule
\multirow{5}{*}{\rotatebox{90}{UniMiB}} 
& 2,3,5    & 26.57 & 23.85 & 26.43 (1.10) & 23.25 (0.23) & 23.88 (0.23) & 24.87 (0.23) & 23.88 (0.11) & 25.16 (1.27) & \textbf{29.88 (0.22)} & \underline{29.64 (0.09)} \\
& 1,3,5    & 38.47 & 39.45 & 36.34 (0.59) & 38.43 (0.29) & 39.22 (0.20) & 39.82 (0.06) & 39.37 (0.12) & 32.89 (1.87) & 44.33 (1.71) & \underline{\textbf{44.52 (1.50)}} \\
& 1,2,5    & 27.46 & 24.13 & 26.14 (1.75) & 23.57 (0.51) & 24.68 (0.16) & 23.59 (0.23) & 24.39 (0.11) & 26.33 (0.42) & \textbf{28.64 (0.93)} & \underline{27.67 (1.34)} \\
& 1,2,3    & 15.29 & 13.66 & 16.83 (0.06) & 11.49 (0.20) & 13.18 (0.40) & 13.04 (0.39) & 13.08 (0.92) & \textbf{17.35 (0.22)} & 13.58 (0.25) & \underline{14.28 (0.26)} \\
& AVG      & 26.95 & 25.27 & 26.43 (0.30) & 24.18 (0.09) & 25.24 (0.17) & 25.33 (0.01) & 25.18 (0.21) & 25.43 (0.20) & \textbf{29.11 (0.78)} & \underline{29.03 (0.75)} \\
\midrule
& AVG all & 49.50 & 50.76 & 51.04 (0.10) & 50.33 (0.09) & 50.86 (0.08) & 51.81 (0.02) & 50.72 (0.09) & 49.20 (0.10) & 51.46 (0.19) & \underline{\textbf{53.58 (0.24)}} \\
\bottomrule
\multicolumn{8}{l}{$\dagger$: Optimization-free, $\psi$: Optimization with gradient descent.}\\
\end{tabular}
\label{tab:7}
}
\end{table*}

\subsubsection{TTA Setting}
We consider two TTA settings following the general TTA protocol~\cite{su2022revisiting, wang2020tent}.
\textbf{(1) \textbf{L}eave-\textbf{O}ne-\textbf{O}ut \textbf{A}daptation (\textbf{LOOA})}. We sequentially leave one domain for the test-time adaptation. The rest of the models are used to pre-train the model on the source domain. \textbf{(2) \textbf{C}ontinual \textbf{T}est-\textbf{T}ime \textbf{A}daptation (\textbf{CTTA})}. The visual description of these two settings is shown in Figure \ref{fig:9}.  We only train our model on one source domain. In the test time, the model continually adapts to several domains with different domain shifts. During the training stage, we only access the training data, and we discard it after training. In the testing phase, following the setting of recent researches \cite{wang2020tent, lim2023ttn,liang2023comprehensive}, we can use an online unlabeled mini-batch to adapt the model and adjust the prediction. For the evaluation metrics, since OPPORTUNITY and UNIMIB-SHAR datasets are class-imbalanced, we exploit both accuracy and macro F1-score as our performance metrics.  All experimental results are obtained through three independent adaptations using different random seeds (1, 2, 3),  implying that the online batch sequence is varied. We report the mean adaptation results with standard deviation.

\subsection{Experimental Results}
The classification results of HAR under LOOA and CTTA settings are shown in Table \ref{tab:4}-\ref{tab:7}. Our proposed OFTTA consistently improves the performance across the three datasets under both the LOOA and CTTA settings. Compared with optimization-free methods, OFTTA achieves substantial performance improvements. When compared with all TTA methods, OFTTA still achieved competitive results. Specifically, compared with baseline ERM, OFTTA achieves a significant accuracy improvement of 3.88\% in LOOA-HAR and 2.77\% in CTTA-HAR. When compared to state-of-the-art TTA methods, OFTTA still outperforms the second-best methods: 0.32\% for LOOA-HAR and 1.41\% for CTTA-HAR. Regarding the macro-F1 score, OFTTA consistently demonstrates a noticeable improvement, signifying the efficacy of our proposed OFTTA in addressing imbalanced HAR datasets. Note that OFTTA may not achieve the best performance in all datasets. For instance, TAST-BN can achieve better results on the OPPO dataset in LOOA-HAR. However, TAST-BN needs specific hyperparameters and gradient optimization. Our proposed OFTTA is hyperparameters-friendly and adapts in an optimization-free manner. More importantly, our proposed method consistently outperforms other state-of-the-art methods in terms of average performance across the three datasets, which shows that our method is adaptive with a better generalization ability Similar can be also observed in macro-F1 score in both TTA settings.

We can make more observations: (1) CTTA is a more challenging setting than LOOA, as we can observe that the baseline model performs poorly in the setting of CTTA. Additionally, the adaptation improvement achieved by TTA methods is lower compared to LOOA. Despite this, our OFTTA still achieves substantial improvements compared with other TTA methods. (2) Several TTA methods suffer from performance degradation. We can take a close look at the results on the UniMiB dataset under the CTTA setting. The results of TTA methods with parameter optimization (TENT, PL, SHOT, SAR, TAST) are worse than baseline ERM. This phenomenon indicates that TTA methods are not risk-free. If the parameters are over-updated, the model may collapse and cause "catastrophic forgetting" issues on the original task. Compared with their methods, our OFTTA adapts the model in an optimization-free manner and mitigates this potential risk.

\section{ABLATION STUDY}
\subsection{Contribution of Each Component in OFTTA}
\begin{figure}[htbp]
  \centering
  \subfigure[UCI]{
    \includegraphics[width=0.3\linewidth]{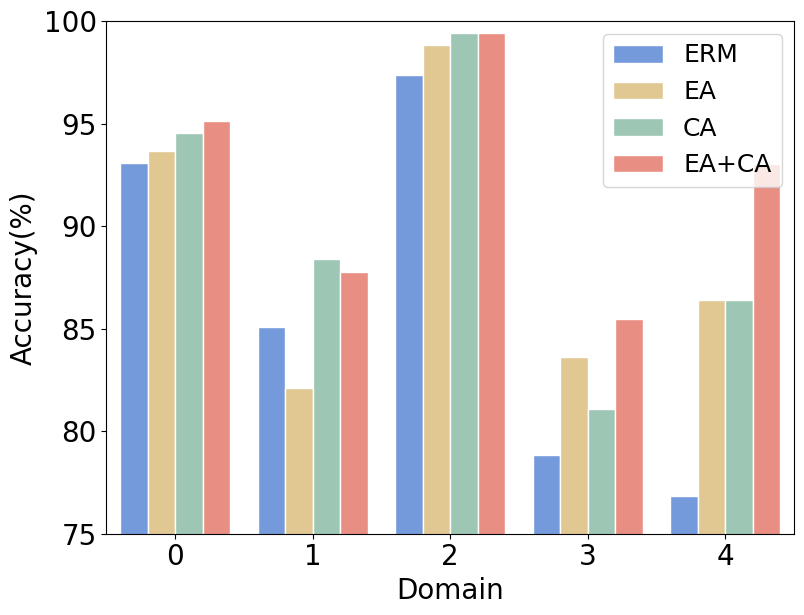}
    \label{fig:subfig10A}
  }
  \subfigure[OPPO]{
    \includegraphics[width=0.3\linewidth]{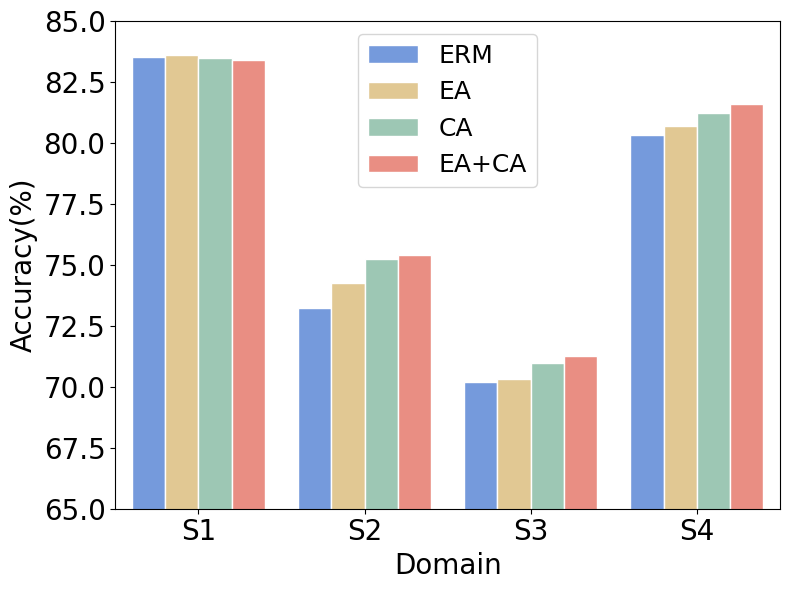}
    \label{fig:subfig10B}
  }
  \subfigure[UniMiB]{
    \includegraphics[width=0.3\linewidth]{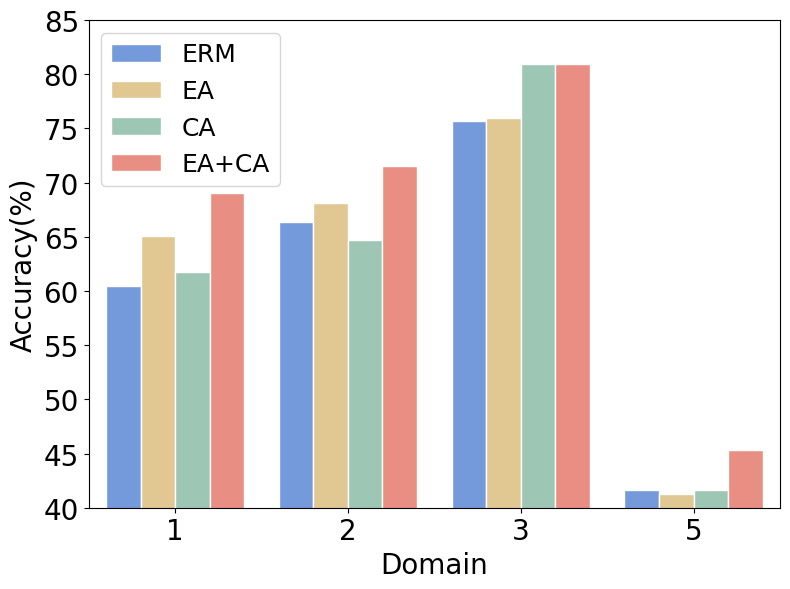}
    \label{fig:subfig10C}
  }
  \vspace{-.2in}
  \caption{The visualization of component effectiveness on three HAR datasets.}
  \vspace{-.1in}
  \label{fig:10}
\end{figure}

OFTTA consists of two key modules for optimization-free adaptation: (1) Encoder Adjustment (EA): We use Exponential Decay Test-time Normalization (EDTN) to adjust the normalization statistics adaptively. (2) Classifier Adjustment (CA): We replace the classifier with a Prototype-based Classification (PC). We choose LOOA as the experiment setting and ablate both of them in Figure \ref{fig:10}. Compared with baseline ERM, introducing EA or CA can both achieve better adaptation results in most cases. We can easily assemble them to construct our OFTTA framework. In almost all cases, OFTTA can strongly improve classification performance. This phenomenon supports our claim. Adjusting the feature extractor can outputs more reliable pseudo labels and further benefit the prototype-based classifier. Taking a counterexample can better understand our claim. We can observe that EDTN encountered a failure in domain 1 of the UCI dataset. The failure further causes the performance of OFTTA to be worse than the adaptation with PC. Since adaptation with EDTN will output worse pseudo labels, the prototype may become biased and lead to misclassification. In short, the results verify that both adjustments are important for adaptation and can work complementarily for better classification results.

\subsection{Parameter Sensitivity Analysis}
We have two hyperparameters in our OFTTA framework and we discuss them separately in this section. 
\subsubsection{Decay factor $\lambda$ in EDTN} The proportion of TBN in the deep layers is inversely proportional to $\lambda$. As we mentioned in section 3.2, we compared our EDTN with other state-of-the-art test-time normalization adjustments:

\begin{enumerate}
    \item \textbf{$\alpha$-BN} \cite{you2021test}: combines CBN and TBN with a pre-defined hyperparameter $\alpha$. The mix ratio is constant in each layer.
    \item \textbf{TTN} (Test-Time Normalization) \cite{lim2023ttn}: adds a post-training stage before the adaptation to mix CBN and TBN based on the score function.
    \item \textbf{TBR} (Test-time Batch Renormalization) \cite{zhao2023delta}: utilizes moving averaged TBN to rectify CBN instead of directly mixing them.
\end{enumerate} 

We first introduce implementations.
For $\alpha$-BN, we can easily set a constant $\alpha$ in our framework.
For TTN, since they did not release the official code, we adopt/modify it from their pseudo code in the paper. TTN needs data augmentation to get the score of the BN parameter. As an alternative, we disturb the sensor input with noise since standard data augmentation may not be suitable for HAR. For TBR, we adopt their official implementation\footnote{\url{https://github.com/bwbwzhao/DELTA}}.

\begin{figure}[t!]
  \centering
  \subfigure[UCI]{
    \includegraphics[width=0.3\linewidth]{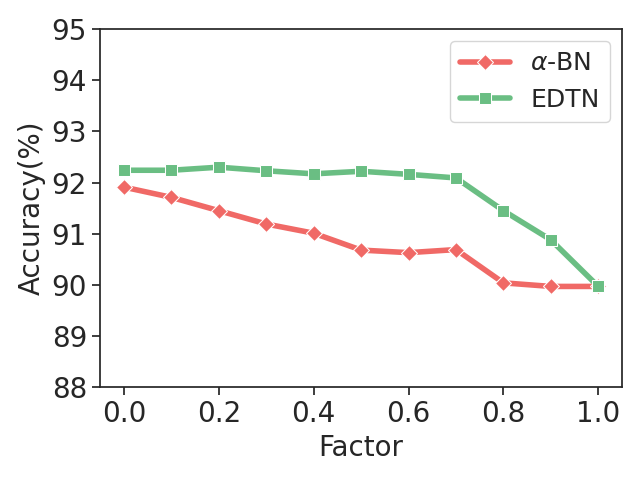}
    \label{fig:11-1}
  }
  \subfigure[OPPO]{
    \includegraphics[width=0.3\linewidth]{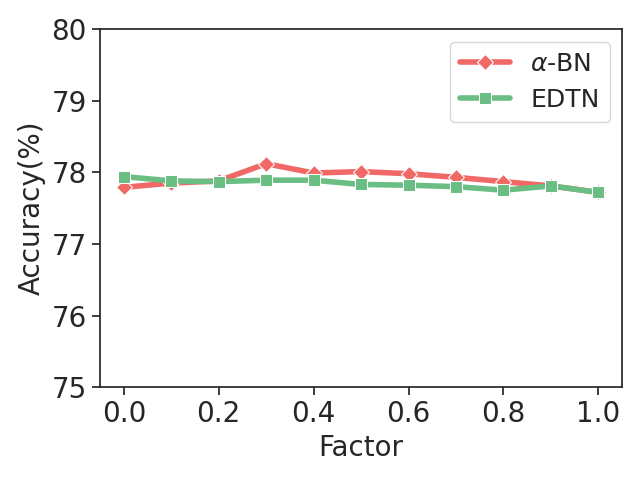}
    \label{fig:11-2}
  }
  \subfigure[UniMiB]{
    \includegraphics[width=0.3\linewidth]{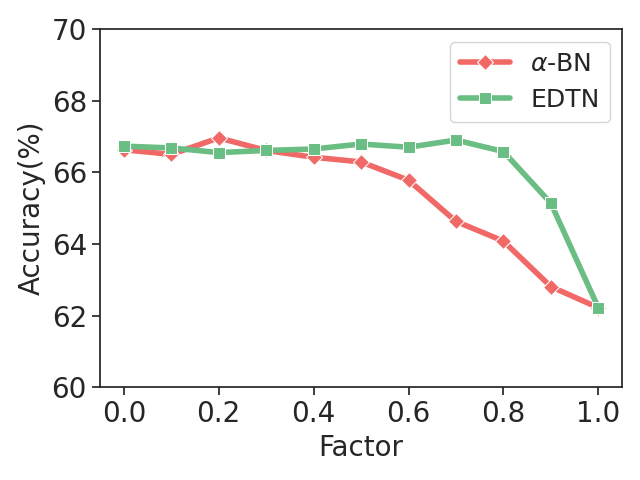}
    \label{fig:11-3}
  }
  \vspace{-.1in}
  \caption{Hyperparameter sensitivity comparison between $\alpha$-BN and EDTN in LOOA.}
  \label{fig:11}
\end{figure}

For fairness, we replace the EDTN module with the above module in our OFTTA framework. The results are illustrated in Figure \ref{fig:11} and Table \ref{tab:8}. The factor in the x-axis represents $\alpha$ in $\alpha$-BN and $\lambda$ in EDTN. We first compare our methods with $\alpha$-BN in Figure \ref{fig:11}. From a macro perspective, EDTN can outperform $\alpha$-BN in most cases. we can also observe that $\alpha$-BN  can be slightly better using some specific $\alpha$ such as OPPO ($\alpha$=0.3) or UniMib ($\alpha$=0.2). However, the results also show that $\alpha$ is sensitive and we can not find one constant $\alpha$ that can perform well on any dataset. We can take a close look at when the factor is 0, EDTN consistently outperforms $\alpha$-BN on three datasets. If the factor is 0, the only difference between the two methods is that we always use CBN in the last BN layer. The results support our claim that the adaptation performance can be benefited from the layer-wise mix of TBN and CBN. We further compared our EDTN with other state-of-the-art BN adjustments in Table \ref{tab:8}. As we mentioned in section 4.2.4, we even do not need to set the actual decay factor $\lambda$. We just set the mix ratio of the top layer and bottom layer, and the decay factor is decided by the number of BN layers in the model. We just need to use TBN in a decreasing manner. Observed that although EDTN may not be optimal on one certain dataset, it outperforms other methods in average performance. The results prove the advantages of EDTN: 
(1) it is parameter insensitive. 
(2) it outperforms other BN adjustment strategies and can improve TTA performance for free. We can also make an important conclusion from the experiment: in test-time adaptation, TBN is suitable for shallow layers and CBN shows its advantage in deep layers.
This observation can also be found in previous studies \cite{lim2023ttn, wang2022revisiting}, and we explain it from the representation structure of modern neural networks in Figure \ref{fig:8}.

\begin{table}[t!]
\caption{Average classification accuracy (\%) with different BN statistics adjustment over three datasets in LOOA. The \textbf{bold} items denote the best of all methods.}
\vspace{-.1in}
\begin{tabular}{ccccccc}
\toprule
Dataset & CBN   & TBN         & $\alpha$-BN($\alpha$=0.3) & TBR   & TTN         & EDTN(ours)     \\ \midrule
UCI     & 89.97 & 91.91       & 91.19                     & 90.03 & 92.03 & \textbf{92.16} \\ 
OPPO    & 77.72 & 77.79       & \textbf{78.12}            & 77.81 & 77.94       & 77.91    \\ 
UniMiB  & 62.22 & 66.63 & 66.61                     & 62.65 & 66.56       & \textbf{66.69} \\ 
AVG     & 76.64 & 78.78       & 78.64                     & 76.83 & 78.84 & \textbf{78.92} \\ \bottomrule

\end{tabular}
\label{tab:8}
\end{table}

\subsubsection{The number of supports $M$.}
\begin{figure}
  \centering
  \subfigure[UCI]{
    \includegraphics[width=0.3\linewidth]{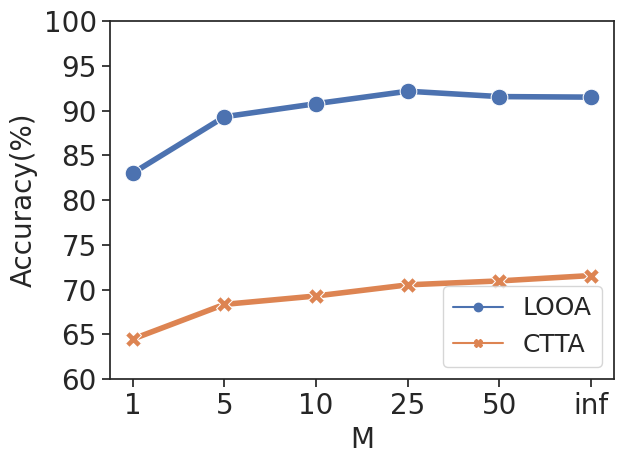}
    \label{fig:subfig12A}
  }
  \subfigure[OPPO]{
    \includegraphics[width=0.3\linewidth]{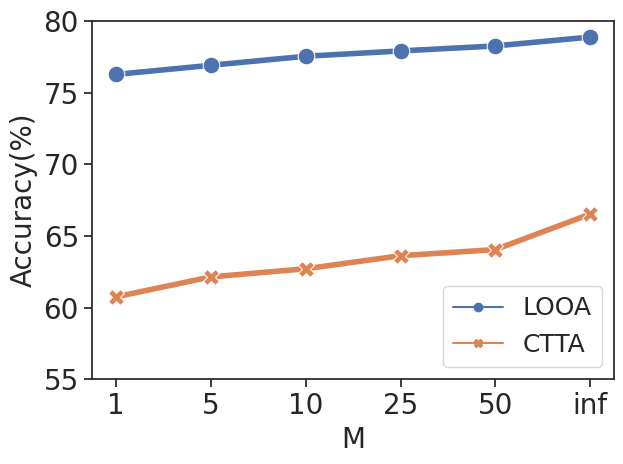}
    \label{fig:subfig12B}
  }
  \subfigure[UniMiB]{
    \includegraphics[width=0.3\linewidth]{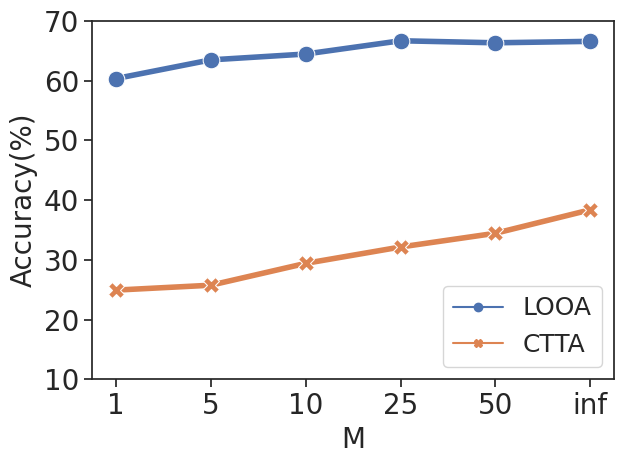}
    \label{fig:subfig12C}
  }
  \vspace{-.1in}
  \caption{Support numbers sensitivity in LOOA and CTTA.}
  \label{fig:12}
\end{figure}

We empirically evaluate the sensitivity of $M$ in the classifier adjustment by setting their values from $\{1, 5, 10, 25, 50, inf\}$, where $inf$ means restoring all samples to adjust the prototype. As is shown in Figure \ref{fig:12}, for LOOA, we observe that the larger $M$ does not always lead to better results since hard samples will pollute the support set and lead to unreliable prototypes. Hence, we empirically set $M$ to 25 and filter the support set based on logit entropy. For CTTA, larger $M$ can bring better performance. This phenomenon is also reasonable as the model adapts each domain sequentially. Some domains can be harder for the pre-train model and lead to high entropy. As a result, the support set does not contain features from these domains and causes worse performance. Specifically,  we calculate the mean entropy of each domain in UCI dataset: $\{0.26, 0.08, 0.20, 0.06, 0.07\}$, corresponding to domain ID: $\{0, 1, 2, 3, 4\}$.  The results support our assumption. It is clear that the features belonging to domain 0 and 2 have evidently higher entropy than other domains, which are more difficult to store in the support set. As a straightforward solution, we store all the features sequentially for the robust prototype. Readers may concern this operation could hurt the efficiency. We empirically verified in section 5.5 that compared with gradient descent, the extra computational cost is quite small.

\subsection{Extending OFTTA with DG Methods}

\begin{figure}
  \centering
  \subfigure[UCI]{
    \includegraphics[width=0.3\linewidth]{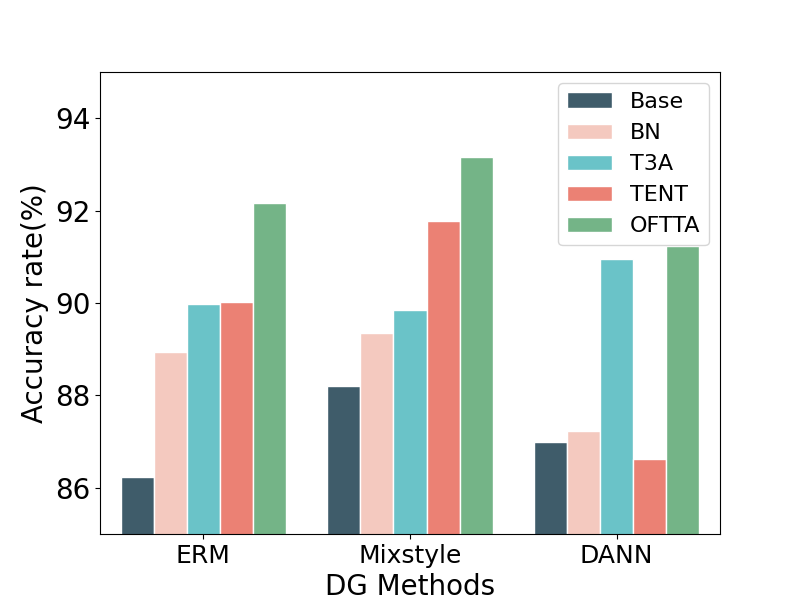}
    \label{fig:subfig13A}
  }
  \subfigure[OPPO]{
    \includegraphics[width=0.3\linewidth]{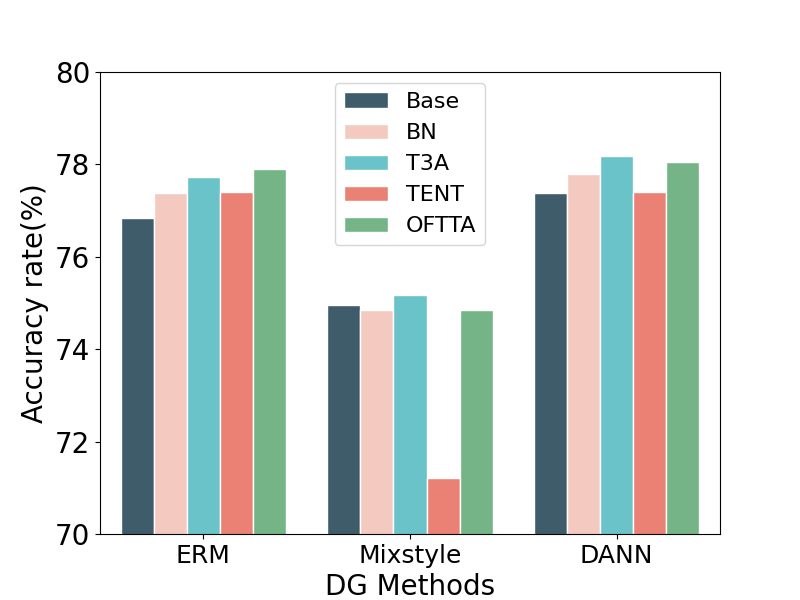}
    \label{fig:subfig13B}
  }
  \subfigure[UniMiB]{
    \includegraphics[width=0.3\linewidth]{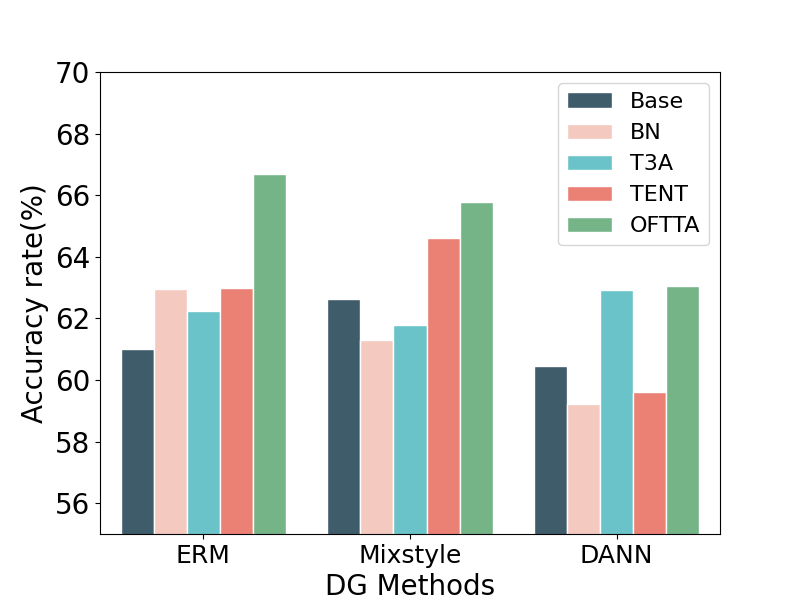}
    \label{fig:subfig13C}
  }
  \vspace{-.2in}
  \caption{Results on three datasets in LOOA. We incorporate domain generalization algorithms into the training phase.}
  \label{fig:13}
\end{figure}

To demonstrate that our proposed OFTTA is a generalizable framework, we incorporate several domain generalization (DG) algorithms into the training phase. Specifically, we adopt two DG algorithms: (1) MixStyle \cite{zhou2021mixstyle}: mix statistics between features from different domains. As we can not access the domain label, we follow the official implementation\footnote{\url{https://github.com/KaiyangZhou/mixstyle-release}} and randomly mix each feature with another feature in the batch. (2)  Domain-Adversarial Neural Networks (DANN) \cite{ganin2016domain} aims to measure domain gaps via an external domain classifier and learn domain-invariant representations. Both methods can be incorporated into training in a $plug-and-play$ manner. As is shown in Figure \ref{fig:13}, our proposed OFTTA demonstrates compatibility with train-time DG methods and further improves the classification performance under domain shifts. We show the average performance over three HAR datasets using DG or TTA in Table \ref{tab:9}. A closed look at the first column can be observed that if we only adopt DG into the training phase, the improvement against distribution shifts may be marginal. If we are only allowed to use DG or TTA, compared with the easiest TTA methods BN (76.43\%), MixStyle only 75.57\% over the three HAR datasets. Since DG can not utilize data from the target domain, it may be unfeasible to generalize a model well or even worse to unseen distributions.
In conclusion, OFTTA is orthogonal to DG  and consistently improves the model performance in the testing phase. Compared with other TTA methods, OFTTA still outperforms them over three HAR datasets.

\begin{table}[t!]
\caption{Classification accuracy (\%) over three datasets using DG/TTA in LOOA. The \textbf{bold} items denote the best of all methods.}
\vspace{-.1in}
\begin{tabular}{cccccc}
\toprule
Train\textbackslash{}Test & Base  & $\text{BN}^\diamond $    & $\text{T3A}^\diamond $   & $\text{TENT}^\diamond $  & $\text{OFTTA}^\diamond $          \\ \midrule
ERM                          & 74.70 & 76.43 & 76.64 & 76.81 & \textbf{78.92} \\ 
$\text{MixStyle}^\Omega  $   & 75.27 & 75.13 & 75.60 & 75.87 & \textbf{77.93} \\ 
$\text{DANN}^\Omega  $       & 74.95 & 74.75 & 77.35 & 74.55 & \textbf{77.45} \\ \bottomrule
\multicolumn{5}{l}{$\Omega$: DG Algorithm,  $\diamond$: TTA Algorithm}
\end{tabular}
\label{tab:9}
\end{table}

\subsection{Backbone Flexibility}
\label{sec:5.4}
To verify the flexibility of OFTTA, we replaced our CNN backbone with two popular networks in HAR: ResNet \cite{he2016deep, huang2022channel, huang2022deep} and DeepConvLSTM \cite{ordonez2016deep}. For ResNet, we just added the residual block to the origin CNN backbone. We use the UCI dataset in LOOA as the experimental setting and compare all TTA methods using these three backbones. The results are shown in Table \ref{tab:10}. Our OFTTA can achieve competitive results and obtain the best average performance across the three HAR datasets. In short, our OFTTA shows great flexibility and can fit diverse backbones.

\begin{table}[t!]
\caption{Average classification accuracy(\%) for different backbones on UCI dataset under LOOA setting. The \textbf{bold} items denote the best of all methods.}
\vspace{-.1in}
\begin{tabular}{cccccccccc}
\toprule
&  $\text{ERM}^\dagger$    & $\text{BN}^\dagger$    & $\text{T3A}^\dagger$              & $\text{TENT}^\psi$  & $\text{PL}^\psi$    & $\text{SHOT}^\psi$           & $\text{SAR}^\psi$ & $\text{TAST-BN}^\psi$        & $\text{OFTTA}^\dagger$ \\ \midrule
CNNs         & 86.24 & 88.93 & 89.97          & 90.03 & 88.73 & 82.78 & 88.86 & 92.14 & \textbf{92.16} \\
ResNet       & 85.82 & 88.29 & \textbf{92.49} & 89.38 & 88.35 & 88.99 & 91.87 & 88.28       &  92.02    \\ 
DeepConvLSTM & 88.96 & 91.90 & 89.22          & 91.99 & 91.29 & 91.22 & 90.77 & 92.04 & \textbf{92.45} \\ 
AVG          & 87.01 & 89.71 & 90.56          & 90.46 & 89.45 & 87.66 & 90.50 & 90.82 & \textbf{92.21} \\ \bottomrule
\multicolumn{8}{l}{$\dagger$: Optimization-free,  $\psi$: Optimization with gradient descent.}\\
\end{tabular}
\label{tab:10}
\end{table}

\subsection{Adaptation with Other Normalization}
\label{sec:5.5}
If we are looking for overall HAR improvement, it is essential to combine TTA techniques with other normalization methods. 
We first emphasize the importance of normalization techniques, which improve models' training stability, optimization efficiency, and generalization ability. As is shown in Table \ref{tab:11}, we show that models without any normalization are the most unstable and had the poorest performance in all three datasets. Recent work also shows that normalization techniques play an important role in modern neural networks \cite{huang2023normalization}. By updating the parameters within these layers, the model can adapt to target distribution efficiently in the test time, which has been widely utilized in prior research \cite{wang2020tent, lim2023ttn, niu2022efficient}. In this work, we mainly focus on adaptation based on the model with batch normalization.  However, we can still make an exploratory experiment using other normalization methods. Inspired by recent work \cite{niu2023towards},  we replaced the original BN with Group Normalization (GN), while keeping the rest of the CNNs unchanged. Different from Batch Normalization (BN), GN does not rely on batch-wise statistics, which cannot be adjusted directly. Hence, If we want to adjust the feature encoder, we need backpropagation to update the parameters of GN. To this end, we proposed two trainable variants of our model: $\text{OFTTA-T}^*$ and OFTTA-T. Both variants use backpropagation to update the parameters in GN. A key difference is that $\text{OFTTA-T}^*$ adopts the same learning rate (LR) for all GN layers,  while OFTTA-T uses a layer-wise exponential decay LR schedule, which is a trainable variant of our proposed EDTN. The results are shown in Table \ref{tab:12}. Although the results of GN-base CNNs are inferior to BN-based CNNs, the trainable version of our model can still improve the adaptation results significantly and achieve competitive results. Compared with other TTA methods, while TAST achieves SOTA results, considering its computation cost (see Table \ref{tab:15}), our proposed OFTTA-T remains practical in real-world scenarios. Moreover, compared with $\text{OFTTA-T}^*$, OFTTA-T achieves noticeable improvements over three datasets, which indicates that our proposed EDTN strategy is effective. Through the lens of the deep model,  freezing parameters of deeper layers and training more shallow layer weights during the testing phase is beneficial.

\begin{table}[]
\caption{ A accuracy (\%) comparison between models equipped with batch normalization, group normalization and no normalization. The \textbf{bold} items denote the best results.}
\begin{tabular}{cccc}
\toprule
       & No norm & Batch norm & Group norm \\ \midrule
UCI    & 74.69 (2.34)   & \textbf{86.24 (0.49)}      & 83.37 (0.44)      \\
OPPO   & 62.37 (1.16)  & \textbf{76.83 (0.12)}      & 76.05 (0.10)      \\
UnimiB & 63.20 (3.71)  & \textbf{74.70 (0.54)}      & 72.95 (0.57)    \\ \bottomrule
\label{tab:11}
\end{tabular}
\end{table}

\begin{table}[]
\caption{Classification accuracy(\%) of TTA methods using different normalization. The \textbf{bold} and {\ul underline} items denote the best and second-best results, respectively.}
\begin{tabular}{cccccccccc}
\toprule &  & ERM   & T3A   & TENT  & SHOT  & SAR   & TAST  & $\text{OFTTA-T}^*$ & OFTTA-T \\ \midrule
\multirow{4}{*}{CNN-BN} & UCI    & 86.24 & 89.97 & 90.03 & 88.78 & 88.86 & 90.23 & \underline{90.68} & \textbf{90.76} \\
                    & OPPO   & 76.83 & 77.73 & 77.40 & \underline{77.78} & 77.31 & \textbf{78.89} & 77.42 & 77.60 \\
                    & UniMiB & 61.02 & 62.23 & 63.00 & 62.98 & 62.81 & 61.70 & \underline{66.95} & \textbf{67.00} \\
                    & AVG    & 74.70 & 76.64 & 76.81 & 76.51 & 76.33 & 76.94 & \underline{78.35} & \textbf{78.45} \\ \midrule
\multirow{4}{*}{CNN-GN} & UCI    & 83.37 & 88.40 & 84.98 & 84.57 & 82.07 & 88.54 & \underline{88.63} & \textbf{88.69}   \\
                    & OPPO   & 76.05 & 74.99 & \textbf{76.09} & \underline{76.05} & 75.78 & 75.39 & 74.91   & 75.23   \\
                    & UniMiB & 59.42 & 61.05 & 59.31 & 59.99 & 59.44 & \textbf{62.13} & 60.71   & \underline{61.21}   \\ 
                    & AVG    & 72.94 & 74.81 & 73.46 & 73.53 & 72.43 & \textbf{75.35} & 74.75   & \underline{75.04}   \\ \bottomrule
\label{tab:12}
\end{tabular}
\end{table}

\section{REAL DEPLOYMENT ON EDGE DEVICES}
Since HAR models are mostly applied to mobile and pervasive computing, to simulate the real-world HAR application, we evaluated our framework from three aspects: target domain adaptation using a single instance (BS=1), performance remaining on the source domain, and real-time inference efficiency.\\
(1) \textbf{Adaptation with BS=1}: We kept the other settings consistent with the main experiment and report the average classification accuracy rate over three HAR datasets with BS=1. For OFTTA, we set the lowest prior rate to 0.6. As the results are shown in Table \ref{tab:13}, in the adaptation task, most optimization-based methods only use TBN and can easily collapse since it is hard to use very few samples(i.e., small batch size) to estimate the statistics accurately. Since our proposed method both considers CBN and TBN in a linear interpolation manner, the classification performance remains consistently competitive. Moreover, compared with T3A, which is a prototype-based adaptation and does not adjust the feature encoder, our proposed OFTTA still achieved better results. Hence, it is effective to incorporate TBN statistics during the adaptation phase even with a single instance.\\
(2) \textbf{Retest on source domain}: We evaluated the classification performance on source domains using the adapted model. The model has been adjusted using the target domain. The second line of Table \ref{tab:13} supports our claim. Due to potential noisy gradients brought by some test samples, the error is accumulated and makes the model lose the prediction ability of the training domains. In contrast, our method alleviates catastrophic forgetting in the adaptation phase in an optimization-free manner and maintains high performance on source domain.

\begin{table}[]
\caption{Classification accuracy (\%) with BS=1. "Source" is the classification performance on source dataset after adaption. The \textbf{bold} and {\ul underline} items denote the best and second-best results, respectively.}
\begin{tabular}{ccccccccccc}
\toprule
       & $\text{ERM}^\dagger$   & $\text{BN}^\dagger$    & $\text{T3A}^\dagger$              & $\text{TENT}^\psi$  & $\text{PL}^\psi$    & $\text{SHOT}^\psi$           & $\text{SAR}^\psi$ & $\text{TAST}^\psi$  & $\text{TAST-BN}^\psi$        & $\text{OFTTA}^\dagger$          \\ \midrule
Target & 61.02  & 47.53 & {\ul 62.54}         & 54.31 & 23.53 & 22.89 & 57.00 &  57.37 & 35.89   & \textbf{63.54} \\
Source & 100.00 & 98.15 & \textbf{99.92} & 37.00 & 0.95  & 2.93  & 97.52 & 94.91       & 31.21   & {\ul 99.50}    \\ \bottomrule
\multicolumn{8}{l}{$\dagger$: Optimization-free,  $\psi$: Optimization with gradient descent.}\\
\label{tab:13}
\end{tabular}
\end{table}

\begin{table}[htbp]
\caption{Parameters of the used server and edge device.}
\vspace{-.1in}
\resizebox{.5\textwidth}{!}{
\begin{tabular}{ccc}
\toprule
Device           & Server             & Raspberry Pi 3+ Model B \\ \midrule
CPU              & i9-10920X @ 3.5GHz & BCM2837B0 @ 1.4Hz       \\
GPU              & GeForce RTX 3090ti & VideoCore VI            \\
RAM              & 32G                & 1G                      \\
Storage          & 1T                 & 64G                     \\
Battery          & CORSAIR AX1600i    & /                       \\
Operating System & Ubuntu 18.04       & Raspbian Buster         \\
DL Framework     & PyTorch 1.10.0     & PyTorch 1.7.1           \\ \bottomrule
\end{tabular}
}
\label{tab:14}
\end{table}

\begin{table}[t!]
\caption{Average memory cache and time cost using TTA algorithms.} 
\vspace{-.1in}
\resizebox{\textwidth}{!}{
\begin{tabular}{cccccccccccc}
\toprule
Criterion                   & Setting & $\text{ERM}^\dagger$   & $\text{BN}^\dagger$    & $\text{T3A}^\dagger$              & $\text{TENT}^\psi$  & $\text{PL}^\psi$    & $\text{SHOT}^\psi$           & $\text{SAR}^\psi$ & $\text{TAST}^\psi$  & $\text{TAST-BN}^\psi$        & $\text{OFTTA}^\dagger$           \\ \midrule
\multirow{2}{*}{Time (ms)}   & LOOA    & 33.73 & 34.07 & 57.96       & 221.41 & 132.83 & 161.15 & 276.83 & 622.98      & 381.96       & 69.23       \\
                            & CTTA    & 35.46 & 40.38 & 57.10       & 229.48 & 148.67 & 201.42 & 264.20 & 609.85      & 386.83       & 68.99       \\ \midrule
\multirow{2}{*}{Memory (MB)} & LOOA    & 0.35  & 0.73  & 4.11        & 11.43  & 8.45  & 13.80  & 13.44  & 7.64        & 14.56        & 4.61        \\
                            & CTTA    & 0.13  & 0.95  & 4.32↑(0.19) & 11.77  & 8.83  & 14.24  & 13.25  & 7.79↑(0.15) & 14.83↑(0.27) & 4.85↑(0.16) \\ \bottomrule
\multicolumn{8}{l}{$\dagger$: Optimization-free,  $\psi$: Optimization with gradient descent.}\\

\end{tabular}
\label{tab:15}
}

\end{table}

\begin{figure*}[t!]
  \centering
  \begin{minipage}[t]{0.355\textwidth}
    \centering
    \includegraphics[width=\textwidth]{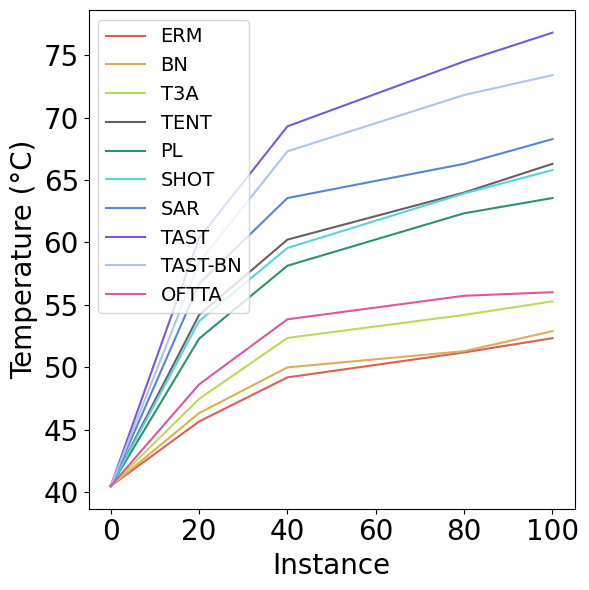}
    \caption{Continuous CPU temperature monitoring of instance-wise adaptation on Raspberry Pi Model 3 B+.}
    \label{fig:14}
  \end{minipage}
  \hspace{0.3cm}
  \begin{minipage}[t]{0.6\textwidth}
    \centering
    \includegraphics[width=\textwidth]{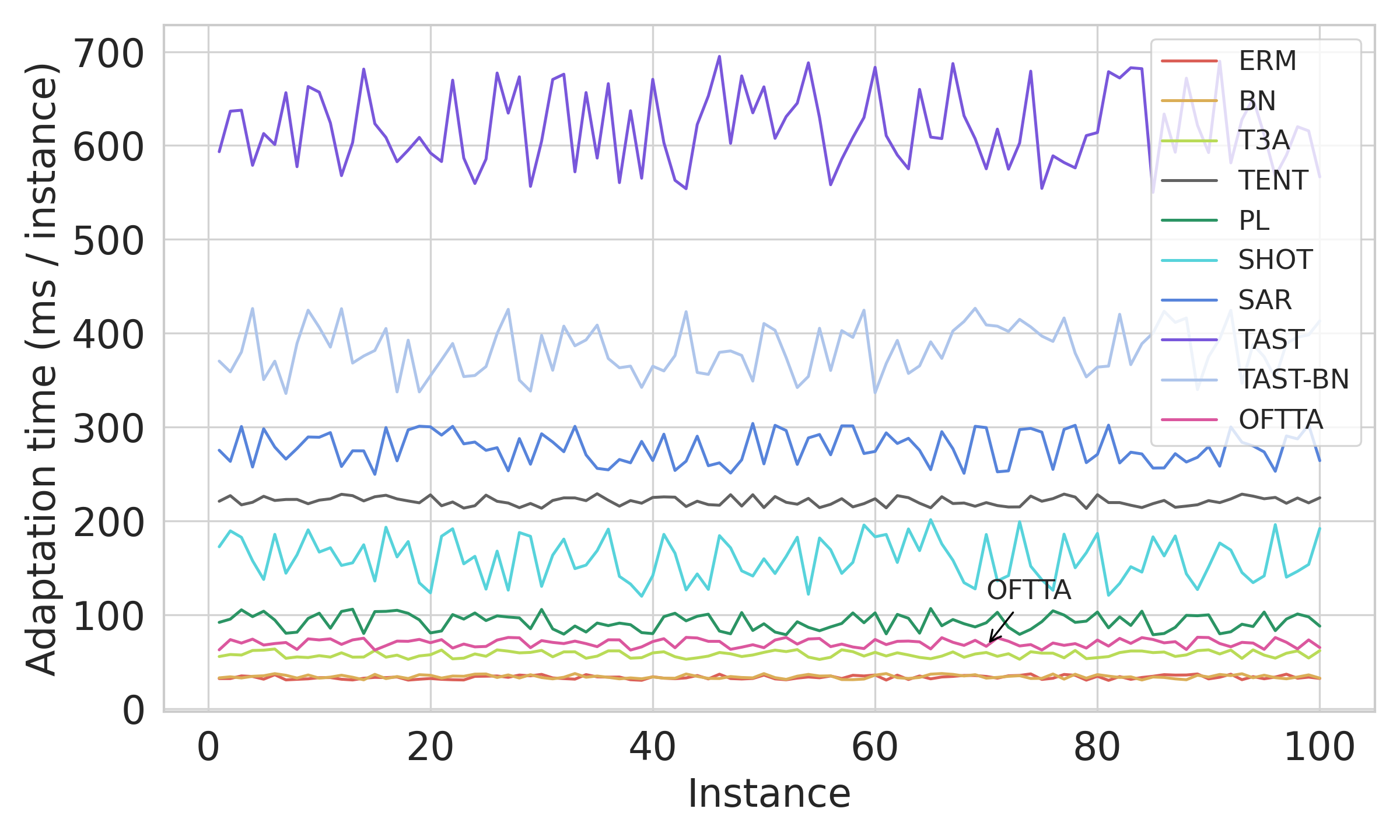}
    \caption{ Elapsed time of instance-wise real-time adaptation (UniMiB dataset) on Raspberry Pi Model 3 B+.}
    \label{fig:15}
  \end{minipage}
\end{figure*}

(3) \textbf{Inference efficiency}: It is necessary to evaluate our framework on resource-constrained edge devices. We evaluate the efficiency of TTA methods on the mobile platform: $Raspberry$ $Pi$ $Model$ $3$ $B+$ with $ARM$ $Cortex$-$A53$. Since the PyTorch library provides excellent support for the Raspberry Pi system, we can easily implement our code on the platform. The main parameter difference between our used server and the edge device is given in Table \ref{tab:14}. Compared with professional servers, the edge device is quite computation-restricted. More importantly, the GPU of almost all mobile devices is poor, which is one key component for the acceleration of the deep learning library. The experimental setup is mostly the same on the server. Specifically, we continuously extract single instances not mini-batch from the UniMiB dataset and send it into the model for real-time adaptation. We track the changes in system memory and elapsed time during the adaptation. We continuously use single instances to adapt the model 100 times. The elapsed time per instance for online adaptation using different TTA methods is visualized in Figure \ref{fig:15}. We can easily observe that optimization-based algorithms (TENT, PL, SHOT, SAR, TAST, TAST-BN) are quite time-consuming compared with optimization-free algorithms(BN, T3A, OFTTA). Moreover, It can be concluded that the model needs more operation time if the optimization objective (e.g. SAR, TAST, TAST-BN) is complex. Compared with optimization-based algorithms, the increased time and memory costs of OFTTA are acceptable. The detailed average time and memory cost are shown in Table \ref{tab:15} in detail.
Taking a close look at prototype-based methods in CTTA,  compared with the optimization cache, the increased memory cost is negligible. In the experiment, to avoid other interference of electronic components like overheating, we test each TTA algorithm independently. Observed that if we use the optimizer-based TTA algorithm, Raspberry Pi can easily overheat and make the adaptation time even longer. Therefore, if we deploy a TTA algorithm on the edge device, it may cause some potential risks raised by the high computational burden of optimization-based methods. To further validate our claim, we monitored the CPU temperature while continuously adapting to 100 instances in one pass. As is shown in Figure \ref{fig:14}, it is not easy to adjust the model in the mobile device during the test phase. Too much computational costs like optimization-based methods will induce a rapid temperature increase in the electronic device. However, the incremental temperature brought by OFTTA is acceptable, which proves the advantage of the optimization-free strategy. Hence, parameter optimization will not only bring potential "catastrophic forgetting" but also lead to high adaptation time and memory cache. Considering our previous observations, we can confidently conclude that our OFTTA can conduct efficient and accurate adaptation on practical HAR tasks.

\section{CONCLUSIONS AND FUTURE WORK}
Distribution shift is a key obstacle hindering HAR applications in the real world. In this paper, we focus on solving covariate shift problems in HAR during the testing phase. We proposed an Optimization-Free Test-Time Adaptation (OFTTA) framework for cross-person HAR by adjusting the feature extractor and classifier simultaneously. For the feature extractor, we propose EDTN to combine the benefits of CBN and TBN. For the classifier, we adopt a support set updated by the pseudo label to obtain the prototype. The adjusted prediction is calculated by the distance between the feature and the prototype. Extensive experiments on three cross-person datasets, two TTA settings, and real edge devices verified that OFTTA is accurate, flexible, and hardware-friendly. Compared with other optimization-free adaption methods, OFTTA achieves substantial performance improvements. When compared with general TTA methods, OFTTA still holds competitive results.

In the future, we plan to pay more attention to continual test-time adaptation (CTTA) since CTTA is a more challenging and realistic setting. For instance, it is important to implement similar improvements on more industry-oriented mobile devices like Jetson Nano. Moreover, in the open world, it may lead to catastrophic outcomes if the model adapts to activities that fall outside the scope of the label space. A simple way to avoid this failure is to detect the out-of-label-space data and refrain from adapting to them. We leave these as future work to conduct stable test-time adaptation for HAR in the dynamic wild world.

\begin{acks}
This research is supported by the National Natural Science Foundation of China (Grant No. 61602540). This work is also in part supported by National Natural Science Foundation of China (Grant No. 62373194). We gratefully acknowledge the support of Center for Computational Science and Engineering at Southern University of Science and Technology for our research.
\end{acks}

\bibliographystyle{ACM-Reference-Format}
\bibliography{OFTTA}

\appendix

\end{document}